\documentclass[11pt]{article}

\usepackage[final]{acl}

\usepackage{times}
\usepackage{latexsym}

\usepackage[T1]{fontenc}

\usepackage[utf8]{inputenc}

\usepackage{microtype}

\usepackage{inconsolata}

\usepackage{graphicx}

\usepackage{tikz}
\usetikzlibrary{arrows.meta, positioning, shapes.geometric, calc, backgrounds, decorations.pathreplacing}

\usepackage{enumitem}

\usepackage{amsmath,amssymb}
\DeclareMathOperator*{\argmin}{arg\,min}

\usepackage{xspace}
\newcommand{\savoir}{\textsc{Savoir}\xspace}

\usepackage{fontawesome5}

\usepackage[ruled,linesnumbered]{algorithm2e}
\SetKwInput{KwInput}{Input}
\SetKwInput{KwOutput}{Output}

\usepackage[most]{tcolorbox}
\tcbuselibrary{listings,skins}

\usepackage{booktabs}
\usepackage{multirow}
\usepackage{colortbl}

\usepackage{listings}
\title{{\savoir}: Learning Social Savoir-Faire via Shapley-based \\ Reward Attribution}

\author{
Xiachong Feng\textsuperscript{1}\thanks{~Equal contribution.},
Yi Jiang\textsuperscript{2}\footnotemark[1],
Xiaocheng Feng\textsuperscript{2}\thanks{~Corresponding author.},
Deyi Yin\textsuperscript{2},
Libo Qin\textsuperscript{3},
Yangfan Ye\textsuperscript{2},\\
\textbf{Lei Huang}\textsuperscript{2},
\textbf{Weitao Ma}\textsuperscript{2},
\textbf{Yuxuan Gu}\textsuperscript{2},
\textbf{Chonghan Qin}\textsuperscript{1},
\textbf{Bing Qin}\textsuperscript{2},
\textbf{Lingpeng Kong}\textsuperscript{1}\footnotemark[2]\\
\textsuperscript{1}The University of Hong Kong \quad
\textsuperscript{2}Harbin Institute of Technology \quad
\textsuperscript{3}Harbin Institute of Technology, Shenzhen\\
\texttt{fengxc@hku.hk, xcfeng@ir.hit.edu.cn, lpk@cs.hku.hk}
}

\begin{document}
\maketitle
\begin{abstract}
Social intelligence, the ability to navigate complex interpersonal interactions, presents a fundamental challenge for language agents. Training such agents via reinforcement learning requires solving the credit assignment problem: determining how individual utterances contribute to multi-turn dialogue outcomes. Existing approaches directly employ language models to distribute episode-level rewards, yielding attributions that are retrospective and lack theoretical grounding. We propose \savoir (\textbf{S}h\textbf{A}pley \textbf{V}alue f\textbf{O}r Soc\textbf{I}al \textbf{R}L), a novel principled framework grounded in cooperative game theory. Our approach combines two complementary principles: \textit{expected utility} shifts evaluation from retrospective attribution to prospective valuation, capturing an utterance's strategic potential for enabling favorable future trajectories; \textit{Shapley values} ensure fair credit distribution with axiomatic guarantees of efficiency, symmetry, and marginality. Experiments on the SOTOPIA benchmark demonstrate that \savoir achieves new state-of-the-art performance across all evaluation settings, with our 7B model matching or exceeding proprietary models including GPT-4o and Claude-3.5-Sonnet. Notably, even large reasoning models consistently underperform, suggesting social intelligence requires qualitatively different capabilities than analytical reasoning.\footnote{Code: \href{https://github.com/jyyyyy0/SAVOIR}{\faGithub\ \texttt{SAVOIR}}}
\end{abstract}

\section{Introduction}

Social intelligence, the capacity to navigate complex interpersonal interactions and achieve social goals, is fundamental to human cognition and increasingly critical for artificial agents \citep{gweon2023socially, lee2024towards}. As large language models (LLMs) become integrated into applications requiring negotiation, collaboration, and persuasion, their ability to exhibit socially intelligent behavior has attracted substantial research attention \citep{zhou2023sotopia, park2023generative, yang2024social}. Yet despite this growing interest, improving the social intelligence of AI systems remains challenging: social interactions are inherently multi-turn, involve competing objectives between participants, and require nuanced understanding of how individual utterances contribute to long-term outcomes \citep{mathur2024advancing, li2024social}.

Recent work has begun addressing these challenges through reinforcement learning (RL) approaches. \citet{wang2024sotopia} propose SOTOPIA-$\pi$, which combines behavior cloning with self-reinforcement on filtered interaction data. More recently, \citet{yu2025sotopia} introduce Sotopia-RL, which refines episode-level feedback into utterance-level rewards by directly prompting an LLM for credit assignment. While Sotopia-RL demonstrates improved performance, its approach exhibits two fundamental limitations. First, the credit assignment mechanism lacks theoretical grounding; the LLM distributes rewards heuristically without principled guarantees of fairness or accuracy. Second, and more critically, the reward model performs \textit{retrospective attribution}: it assigns credit based on what an utterance contributed to the observed outcome, rather than evaluating its \textit{strategic value} for enabling favorable future trajectories. This distinction matters because socially intelligent behavior often involves utterances whose immediate contribution appears minimal but whose strategic positioning unlocks subsequent success.

To address these limitations, we propose \savoir (\textbf{S}h\textbf{A}pley \textbf{V}alue f\textbf{O}r Soc\textbf{I}al \textbf{R}L), a theoretically grounded framework that reconceptualizes credit assignment through two complementary principles from game theory. First, we adopt \textbf{expected utility} to shift the evaluation focus from retrospective attribution to prospective valuation. Rather than asking ``what did this utterance contribute to the final outcome?'', we ask ``what is the expected value of future interactions given this utterance?'' By computing expected outcomes over all possible partner responses and subsequent dialogue trajectories, we capture an utterance's \textit{strategic potential}, its capacity to establish favorable conditions for future success. Second, we employ \textbf{Shapley values} from cooperative game theory to distribute this strategic value fairly across utterances. The Shapley value provides the unique attribution method satisfying efficiency, symmetry, and marginality axioms \citep{lundberg2017unified}, ensuring that utterances receive credit proportional to their true marginal contribution across all possible orderings. Together, these principles transform credit assignment from a heuristic into a principled computation: expected utility defines \textit{what} we measure (forward-looking strategic value), while Shapley values determine \textit{how} we distribute it (fair, axiomatic attribution).

We evaluate \savoir comprehensively on the SOTOPIA benchmark \citep{zhou2023sotopia}, comparing against proprietary LLMs, large reasoning models, and state-of-the-art social intelligence methods. Experiments demonstrate that \savoir achieves new state-of-the-art performance across all evaluation settings: on SOTOPIA-Hard with GPT-4o as partner, the most challenging setting, \savoir obtains a Goal score of 7.18, improving over the strongest baseline by 7.5\%. Notably, our 7B model matches or exceeds proprietary LLMs including GPT-4o and Claude-3.5-Sonnet, while large reasoning models (OpenAI-o1, Gemini-2.5-Pro, DeepSeek-R1) consistently underperform despite their strong analytical capabilities, suggesting that social intelligence requires qualitatively different skills. Human evaluation with expert annotators further validates that \savoir produces more strategic responses and that its reward model better captures nuanced credit assignment.

Our contributions are threefold:
\begin{itemize}[leftmargin=*, nosep]
    \item We propose \savoir, a theoretically grounded framework for social RL that combines expected utility for prospective valuation with Shapley values for fair credit attribution.
    \item We demonstrate state-of-the-art performance on SOTOPIA benchmarks, with a 7B model matching proprietary LLMs and revealing that reasoning models underperform on social tasks.
    \item We provide extensive analysis including human evaluation, ablation studies, and case studies that validate the effectiveness of our principled credit assignment approach.
\end{itemize}
\section{Preliminaries}
\label{sec:preliminaries}

This section provides the foundational concepts for our work. We first present the training pipeline for social agents (\S\ref{subsec:pipeline}), then formalize the social interaction task (\S\ref{subsec:formulation}), and finally describe the evaluation framework (\S\ref{subsec:evaluation}).

\subsection{Training Pipeline Overview}
\label{subsec:pipeline}

\begin{figure}[t]
\centering
\resizebox{\linewidth}{!}{%
\begin{tikzpicture}[
    >={Stealth[length=2mm]},
    box/.style={
        rectangle,
        rounded corners=4pt,
        draw=#1!70!black,
        fill=#1!12,
        thick,
        minimum height=1.1cm,
        minimum width=2.4cm,
        align=center,
        font=\small\sffamily
    },
    arrow/.style={->, thick, >=Stealth, color=gray!70!black}
]

\node[box=blue] (selfplay) at (0, 0) {GPT\\Self-Play};
\node[box=blue] (episodes) at (0, -1.8) {Social\\Episodes $\{\tau_i\}$};

\node[box=orange] (attribution) at (4.2, 0) {Reward\\Attribution};
\node[box=orange] (aggregation) at (4.2, -1.8) {Multi-Dim\\Aggregation};

\node[box=purple] (rm) at (7.4, -1.8) {Reward\\Model};

\node[box=teal] (sft) at (10.6, 0) {SFT\\Warm-up};
\node[box=teal] (rl) at (10.6, -1.8) {Online RL\\(GRPO)};

\draw[arrow] (selfplay) -- (episodes);
\draw[arrow] (attribution) -- (aggregation);
\draw[arrow] (sft) -- (rl);

\draw[arrow] (episodes) -- node[above, font=\scriptsize, text=black] {$\tau, G$} (attribution);
\draw[arrow] (aggregation) -- node[above, font=\scriptsize, text=black] {$r_t$} (rm);
\draw[arrow] (rm) -- node[above, font=\scriptsize, text=black] {$R_\theta$} (rl);

\draw[arrow] (episodes.east) to[out=15, in=165] node[below, pos=0.75, font=\scriptsize, text=black] {SFT data} (sft.west);

\begin{scope}[on background layer]
    \fill[blue!8, rounded corners=6pt] (-1.5, 0.9) rectangle (1.5, -2.7);
    \fill[orange!8, rounded corners=6pt] (2.7, 0.9) rectangle (8.9, -2.7);
    \fill[teal!8, rounded corners=6pt] (9.1, 0.9) rectangle (12.1, -2.7);
\end{scope}

\node[font=\footnotesize\sffamily\bfseries, text=blue!70!black] at (0, 1.3) {Stage 1};
\node[font=\footnotesize\sffamily\bfseries, text=orange!70!black] at (5.8, 1.3) {Stage 2};
\node[font=\footnotesize\sffamily\bfseries, text=teal!70!black] at (10.6, 1.3) {Stage 3};

\node[font=\scriptsize\sffamily, text=blue!60!black] at (0, 0.95) {Data Collection};
\node[font=\scriptsize\sffamily, text=orange!60!black] at (5.8, 0.95) {Reward Modeling};
\node[font=\scriptsize\sffamily, text=teal!60!black] at (10.6, 0.95) {Policy Training};

\end{tikzpicture}
}
\caption{Overview of the social agent training pipeline. \textbf{Stage 1}: Collect social interaction episodes through LLM self-play. \textbf{Stage 2}: Design utterance-level, multi-dimensional rewards through attribution and aggregation. \textbf{Stage 3}: Train the policy via supervised fine-tuning followed by online reinforcement learning with the learned reward model.}
\label{fig:pipeline}
\end{figure}
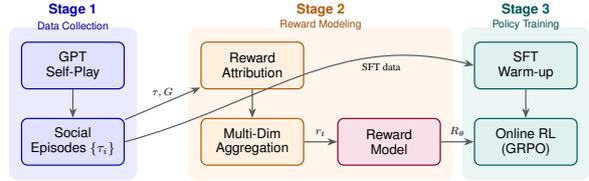

Figure~\ref{fig:pipeline} illustrates the standard training pipeline for social agents. The process consists of three stages: (1) \textbf{data collection}, where LLM agents engage in self-play to generate social interaction episodes; (2) \textbf{reward modeling}, where episode-level outcomes are attributed to individual utterances and aggregated across multiple evaluation dimensions; and (3) \textbf{policy training}, where the agent is first warmed up through supervised fine-tuning and then optimized via online reinforcement learning using the trained reward model.

\begin{figure*}[t]
    \centering
    \resizebox{\textwidth}{!}{%
    \begin{tikzpicture}[
        >={Stealth[length=3mm, width=2mm]},
        mainbox/.style={
            rectangle,
            rounded corners=6pt,
            draw=#1!60!black,
            fill=#1!15,
            line width=1.2pt,
            minimum height=1.4cm,
            minimum width=3.0cm,
            align=center,
            font=\sffamily
        },
        smallbox/.style={
            rectangle,
            rounded corners=4pt,
            draw=#1!50!black,
            fill=#1!12,
            line width=0.8pt,
            minimum height=0.9cm,
            minimum width=2.2cm,
            align=center,
            font=\small\sffamily
        },
        stepnum/.style={
            circle,
            fill=#1!70!black,
            text=white,
            font=\sffamily\bfseries\small,
            minimum size=0.6cm,
            inner sep=0pt
        },
        arrow/.style={->, line width=1.5pt, color=#1!60!black},
        grouplabel/.style={font=\sffamily\bfseries, text=#1!70!black},
        sublabel/.style={font=\scriptsize\sffamily, text=gray!60!black}
    ]

    \node[mainbox=blue] (input) at (0, 0) {\textbf{Social Dialogue}\\$\tau$, $N\!=\!\{a_1, \ldots, a_n\}$};
    \node[stepnum=blue, above left=-0.2cm and -0.2cm of input] {1};

    \node[mainbox=orange] (coalition) at (4.0, 0) {\textbf{Coalition Sampling}\\$\mathcal{C} \subseteq 2^N$};
    \node[stepnum=orange, above left=-0.2cm and -0.2cm of coalition] {2};
    \node[sublabel, below=0.1cm of coalition] {KernelSHAP weighting};

    \begin{scope}[on background layer]
    \fill[teal!10, rounded corners=8pt] (6.4, -1.6) rectangle (12.4, 1.6);
    \end{scope}
    \node[grouplabel=teal] at (9.4, 1.25) {Expected Utility};

    \node[smallbox=purple] (history) at (7.8, 0.5) {\textbf{History}\\$H(S)$};
    \node[smallbox=purple] (rollout) at (7.8, -0.6) {\textbf{Rollout}\\$\pi_A, \pi_B$};
    \node[smallbox=teal] (value) at (10.8, 0.5) {\textbf{Value}\\$v(S)$};
    \node[smallbox=teal] (weight) at (10.8, -0.6) {\textbf{Weight}\\$w_S$};
    \node[stepnum=teal, above left=-0.2cm and -0.2cm of history] {3};

    \begin{scope}[on background layer]
    \fill[red!8, rounded corners=8pt] (13.0, -1.6) rectangle (18.6, 1.6);
    \end{scope}
    \node[grouplabel=red] at (15.8, 1.25) {Credit Assignment};

    \node[smallbox=red] (regression) at (14.4, 0) {\textbf{Regression}\\$\boldsymbol{\phi}$};
    \node[smallbox=red] (normalize) at (17.2, 0) {\textbf{Normalize}\\$\hat{\phi}_i \in [0,10]$};
    \node[stepnum=red, above left=-0.2cm and -0.2cm of regression] {4};

    \draw[arrow=blue] (input) -- (coalition);
    \draw[arrow=orange] (coalition) -- (history.west);
    \draw[arrow=orange] (coalition) -- (rollout.west);
    \draw[arrow=purple] (history.east) -- (value.west);
    \draw[arrow=purple] (rollout.east) -- (weight.west);
    \draw[arrow=teal] (value.east) -- ++(0.6,0) |- (regression.west);
    \draw[arrow=teal] (weight.east) -- ++(0.6,0) |- (regression.west);
    \draw[arrow=red] (regression) -- (normalize);

    \end{tikzpicture}%
    }
    \caption{Overview of the \savoir framework. \textbf{Step 1}: Input social dialogue $\tau$ with agent utterances $N = \{a_1, \ldots, a_n\}$. \textbf{Step 2}: Sample coalitions $\mathcal{C}$ using KernelSHAP weighting. \textbf{Step 3}: For each coalition $S$, reconstruct history $H(S)$, perform rollouts to compute value $v(S)$, and derive SHAP weight $w_S$. \textbf{Step 4}: Solve weighted regression to obtain Shapley values $\boldsymbol{\phi}$, then normalize to $[0,10]$.}
    \label{fig:method_overview}
\end{figure*}
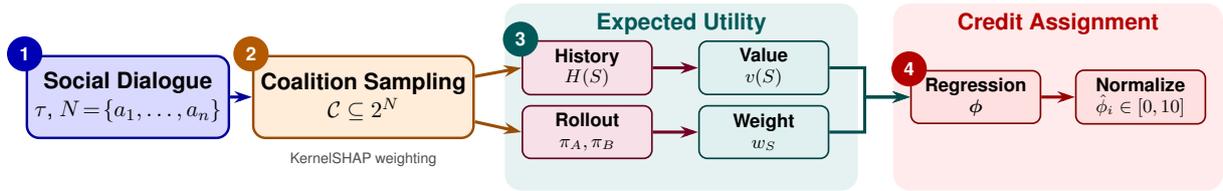

\subsection{Task Formulation}
\label{subsec:formulation}

Social interaction can be formalized as a partially observable Markov decision process (POMDP), defined by the tuple $\langle \mathcal{S}, \mathcal{A}, \mathcal{O}, \mathcal{T}, Z, R \rangle$, where $\mathcal{S}$ denotes the state space, $\mathcal{A}$ the action space, $\mathcal{O}$ the observation space, $\mathcal{T}: \mathcal{S} \times \mathcal{A} \rightarrow \mathcal{S}$ the transition function, $Z: \mathcal{S} \rightarrow \mathcal{O}$ the observation function, and $R: \mathcal{S} \times \mathcal{A} \rightarrow \mathbb{R}$ the reward function. A social episode with $T$ turns is represented as $\tau = (o_0, a_0, o_1, a_1, \ldots, o_T)$, where $o_t \in \mathcal{O}$ is the dialogue history observed at turn $t$ and $a_t \in \mathcal{A}$ is the utterance generated by the agent. Given a private goal $g$, the agent samples actions according to its policy $\pi_\theta(\cdot \mid o_t, g)$.

\paragraph{Reward Modeling.}
The central challenge lies in designing effective reward signals. Given an episode $\tau$ and goal $g$, an LLM-based evaluator provides an episode-level score $G = f(\tau, g) \in \mathbb{R}$. However, episode-level rewards offer only coarse supervision. To obtain fine-grained signals, we attribute the outcome to individual utterances: $r_t = G \cdot \mathcal{A}(a_t, \tau)$, where $\mathcal{A}(a_t, \tau) \in [0, 1]$ represents the contribution of utterance $a_t$ to the episode outcome, estimated by an LLM with access to the full dialogue context. Furthermore, social interactions are inherently multi-dimensional. Beyond goal completion, utterances may contribute to relationship building, knowledge exchange, or other social objectives. We aggregate rewards across $D$ dimensions: $r_t = \frac{1}{D} \sum_{d=1}^{D} w_d \cdot \tilde{r}_{t,d}$, where $\tilde{r}_{t,d}$ is the normalized reward for dimension $d$ and $w_d$ is its corresponding weight.

\subsection{SOTOPIA Evaluation Suite}
\label{subsec:evaluation}

SOTOPIA \citep{zhou2023sotopia} provides an open-ended environment for evaluating social intelligence. Agents role-play through social scenarios, including negotiation, persuasion, collaboration, and accommodation, each with private goals hidden from the interaction partner. The environment evaluates agent performance along seven dimensions: \textbf{Goal Completion} (\textsc{Goal}), the primary metric measuring task success; \textbf{Believability} (\textsc{Bel}), consistency with the assigned persona; \textbf{Relationship} (\textsc{Rel}), maintenance of positive rapport; \textbf{Knowledge} (\textsc{Kno}), appropriate information exchange; \textbf{Secret} (\textsc{Sec}), protection of private information; \textbf{Social Rules} (\textsc{Soc}), adherence to social norms; and \textbf{Financial} (\textsc{Fin}), material outcomes when applicable.
This multi-dimensional evaluation enables comprehensive assessment of social intelligence, capturing both outcome-oriented success and process-oriented interaction quality.

\section{Method}
\label{sec:method}

Building upon the preliminaries, we present \savoir (\textbf{S}h\textbf{A}pley \textbf{V}alue f\textbf{O}r Soc\textbf{I}al \textbf{R}L), a principled framework for computing utterance-level rewards in social interactions. Just as \textit{savoir-faire}, the French term for social grace, captures the art of knowing how to act appropriately in social situations, \savoir teaches language agents this skill through game-theoretic reward attribution. Our approach leverages two fundamental concepts: \textit{expected utility} for evaluating strategic potential and \textit{Shapley value} for fair credit assignment. We first provide an overview of our framework (\S\ref{subsec:overview}), then detail the expected utility formulation (\S\ref{subsec:expected_utility}), the Shapley value-based credit assignment (\S\ref{subsec:shapley}), and the efficient computation via KernelSHAP (\S\ref{subsec:kernelshap}). Finally, we describe the reward model training procedure (\S\ref{subsec:rm_training}).

\subsection{Framework Overview}
\label{subsec:overview}

The core challenge in reward modeling for social interactions lies in attributing episode-level outcomes to individual utterances. Existing approaches either use coarse episode-level rewards or rely on heuristic credit assignment, both of which fail to capture the strategic nature of social dialogue. We address this challenge by formulating reward computation as a cooperative game where each utterance is a player contributing to the collective outcome.

Figure~\ref{fig:method_overview} illustrates the \savoir framework. Given a dialogue $\tau$ containing $n$ utterances from the target agent, denoted as $N = \{a_1, \ldots, a_n\}$, our goal is to compute a reward $\phi_i$ for each utterance $a_i$ that reflects its strategic contribution. The framework operates in three stages: (1) sampling coalitions of utterances, (2) evaluating the expected utility of each coalition through rollouts, and (3) computing Shapley values to distribute credit.

\subsection{Expected Utility for Strategic Evaluation}
\label{subsec:expected_utility}

\paragraph{Motivation.}
Traditional reward attribution methods evaluate utterances based on their historical contribution to the final outcome. However, in strategic social interactions, the value of an utterance lies not only in what has been achieved but also in what \textit{can be achieved} from the current state. For instance, a well-crafted proposal may open pathways to favorable outcomes that are not immediately apparent. To capture this forward-looking perspective, we adopt expected utility theory from decision science, which evaluates actions based on their anticipated future value.

\paragraph{Formulation.}
We define a value function $v: 2^N \rightarrow \mathbb{R}$ that maps any subset (coalition) of utterances $S \subseteq N$ to a scalar value representing its strategic worth. Formally, for a coalition $S$, the value function is defined as:
\begin{equation}
    v(S) = \mathbb{E}_{\tau' \sim \mathcal{R}(H(S))} \left[ U(\tau') \right],
    \label{eq:value_function}
\end{equation}
where $H(S)$ denotes the reconstructed dialogue history containing only utterances in $S$ along with their corresponding partner responses, $\mathcal{R}(H(S))$ represents the distribution over future dialogue trajectories starting from state $H(S)$, and $U(\tau')$ is the utility of a complete trajectory.

\paragraph{Future Rollout.}
To compute the expectation in Eq.~\ref{eq:value_function}, we perform Monte Carlo simulation. Starting from the reconstructed history $H(S)$, we conduct $J$ complete dialogues using the agent policy $\pi_A$ and a partner policy $\pi_B$:
\begin{equation}
    v(S) = \frac{1}{J} \sum_{j=1}^{J} U(\tau_j),
    \label{eq:monte_carlo}
\end{equation}
where each $\tau_j$ is a complete trajectory obtained by alternating between $\pi_A$ and $\pi_B$ until the dialogue terminates.

\paragraph{Utility Function.}
The utility $U(\tau)$ of a trajectory is computed using the SOTOPIA evaluation framework, which provides scores across multiple dimensions. We aggregate these dimensions using a weighted combination following \citet{yu2025sotopia}:
\begin{equation}
    U(\tau) = \sum_{d=1}^{D} w_d \cdot G_d(\tau),
    \label{eq:utility}
\end{equation}
where $G_d(\tau)$ is the score for dimension $d$ and $w_d$ is its corresponding weight. This formulation allows flexible emphasis on different social objectives such as goal completion, relationship maintenance, or norm adherence.

\subsection{Shapley Value for Credit Assignment}
\label{subsec:shapley}

\paragraph{Motivation.}
With the value function defined, we now face the credit assignment problem: how to fairly distribute the total value among individual utterances? Consider a negotiation where multiple utterances collectively lead to a successful agreement. Some utterances may establish rapport, others may introduce key proposals, and still others may handle objections. A principled attribution method should recognize the unique contribution of each utterance, accounting for its synergistic effects with other utterances.

\paragraph{Formulation.}
The Shapley value from cooperative game theory provides an axiomatic solution to this problem. For a cooperative game defined by a player set $N$ and a value function $v$, the Shapley value $\phi_i$ of player $i$ is the weighted average of its marginal contributions across all orderings:
\begin{equation}
    \small
    \phi_i = \sum_{S \subseteq N \setminus \{i\}} \frac{|S|!(n-|S|-1)!}{n!} \left[ v(S \cup \{i\}) - v(S) \right].
    \label{eq:shapley}
\end{equation}

The term $v(S \cup \{i\}) - v(S)$ represents the marginal contribution of utterance $a_i$ to coalition $S$, and the coefficient ensures that each ordering is weighted equally. The Shapley value satisfies four desirable properties: \textit{efficiency} (the values sum to $v(N) - v(\emptyset)$), \textit{symmetry} (identical contributions receive identical values), \textit{null player} (zero contribution implies zero value), and \textit{additivity} (values are additive across games).

\paragraph{Interpretation.}
In our context, the Shapley value $\phi_i$ quantifies the average marginal contribution of utterance $a_i$ to the expected future utility. A high Shapley value indicates that the utterance consistently improves outcomes when added to various coalitions, suggesting strong strategic value. Conversely, a low or negative value indicates that the utterance may be redundant or even detrimental.

\begin{figure}[t]
\centering
\resizebox{\linewidth}{!}{%
\begin{tikzpicture}[
    utt/.style={rectangle, rounded corners=3pt, draw=#1!60!black, fill=#1!20, minimum width=0.9cm, minimum height=0.6cm, font=\small\sffamily\bfseries},
    empty/.style={rectangle, rounded corners=3pt, draw=gray!40, fill=gray!5, minimum width=0.9cm, minimum height=0.6cm, font=\small\sffamily, text=gray!50},
    valuebox/.style={rectangle, rounded corners=2pt, fill=#1!15, draw=#1!50!black, minimum width=1.2cm, font=\small},
    arrowline/.style={->, thick, gray!60},
    orderlabel/.style={font=\scriptsize\sffamily, text=gray!60!black}
]

\node[font=\sffamily\bfseries, text=gray!70!black] at (4.5, 2.5) {Computing $\phi_{a_2}$: Marginal Contributions across Permutations};

\node[orderlabel] at (0.1, 1.8) {$\pi_1$:};
\node[utt=blue] at (1.4, 1.8) {$a_2$};
\node[utt=orange] at (2.1, 1.8) {$a_1$};
\node[utt=red] at (2.8, 1.8) {$a_3$};
\node[valuebox=green] (v1) at (4.8, 1.8) {$+1.2$};
\node[font=\scriptsize, text=gray, right=0.1cm of v1] {$v(\{a_2\})\!-\!v(\emptyset)$};

\node[orderlabel] at (0.1, 1.1) {$\pi_2$:};
\node[utt=blue] at (1.4, 1.1) {$a_2$};
\node[utt=red] at (2.1, 1.1) {$a_3$};
\node[utt=orange] at (2.8, 1.1) {$a_1$};
\node[valuebox=green] (v2) at (4.8, 1.1) {$+1.2$};
\node[font=\scriptsize, text=gray, right=0.1cm of v2] {$v(\{a_2\})\!-\!v(\emptyset)$};

\node[orderlabel] at (0.1, 0.4) {$\pi_3$:};
\node[utt=orange] at (1.4, 0.4) {$a_1$};
\node[utt=blue] at (2.1, 0.4) {$a_2$};
\node[utt=red] at (2.8, 0.4) {$a_3$};
\node[valuebox=green] (v3) at (4.8, 0.4) {$+0.8$};
\node[font=\scriptsize, text=gray, right=0.1cm of v3] {$v(\{a_1,a_2\})\!-\!v(\{a_1\})$};

\node[orderlabel] at (0.1, -0.3) {$\pi_4$:};
\node[utt=red] at (1.4, -0.3) {$a_3$};
\node[utt=blue] at (2.1, -0.3) {$a_2$};
\node[utt=orange] at (2.8, -0.3) {$a_1$};
\node[valuebox=green] (v4) at (4.8, -0.3) {$+1.0$};
\node[font=\scriptsize, text=gray, right=0.1cm of v4] {$v(\{a_3,a_2\})\!-\!v(\{a_3\})$};

\node[orderlabel] at (0.1, -1.0) {$\pi_5$:};
\node[utt=orange] at (1.4, -1.0) {$a_1$};
\node[utt=red] at (2.1, -1.0) {$a_3$};
\node[utt=blue] at (2.8, -1.0) {$a_2$};
\node[valuebox=green] (v5) at (4.8, -1.0) {$+0.6$};
\node[font=\scriptsize, text=gray, right=0.1cm of v5] {$v(N)\!-\!v(\{a_1,a_3\})$};

\node[orderlabel] at (0.1, -1.7) {$\pi_6$:};
\node[utt=red] at (1.4, -1.7) {$a_3$};
\node[utt=orange] at (2.1, -1.7) {$a_1$};
\node[utt=blue] at (2.8, -1.7) {$a_2$};
\node[valuebox=green] (v6) at (4.8, -1.7) {$+0.6$};
\node[font=\scriptsize, text=gray, right=0.1cm of v6] {$v(N)\!-\!v(\{a_1,a_3\})$};

\draw[thick, dashed, gray!50] (0, -2.2) -- (9, -2.2);
\node[rectangle, rounded corners=4pt, fill=teal!15, draw=teal!60!black, thick, font=\sffamily] at (4.5, -2.75) {$\phi_{a_2} = \frac{1.2 + 1.2 + 0.8 + 1.0 + 0.6 + 0.6}{6} = \mathbf{0.9}$};

\end{tikzpicture}%
}
\caption{Shapley value computation for $a_2$. For each of the $n!=6$ permutations, we compute $a_2$'s marginal contribution when it joins. The Shapley value is the average across all permutations. See Appendix~\ref{sec:appendix_shapley} for detailed explanation.}
\label{fig:shapley_intuition}
\end{figure}

\subsection{Efficient Computation via KernelSHAP}
\label{subsec:kernelshap}

\paragraph{Computational Challenge.}
Direct computation of Shapley values requires evaluating $v(S)$ for all $2^n$ subsets, which is computationally prohibitive for dialogues with many utterances. Moreover, each evaluation of $v(S)$ requires $J$ rollout simulations, further compounding the cost.

\paragraph{KernelSHAP Algorithm.}
To address this challenge, we employ KernelSHAP \citep{lundberg2017unified}, which reformulates Shapley value computation as a weighted linear regression. The insight is that Shapley values can be obtained by solving:
\begin{equation}
    \boldsymbol{\phi}^* = \argmin_{\boldsymbol{\phi}} \sum_{k=1}^{K} w_k \left( v(S_k) - \sum_{i=1}^{n} \phi_i \cdot z_{ki} \right)^2,
    \label{eq:kernelshap}
\end{equation}
where $\{S_k\}_{k=1}^K$ are sampled coalitions, $z_{ki} \in \{0, 1\}$ indicates whether utterance $a_i$ is in coalition $S_k$, and $w_k$ is the SHAP kernel weight:
\begin{equation}
    w_k = \frac{n-1}{\binom{n}{|S_k|} \cdot |S_k| \cdot (n - |S_k|)}.
    \label{eq:shap_kernel}
\end{equation}

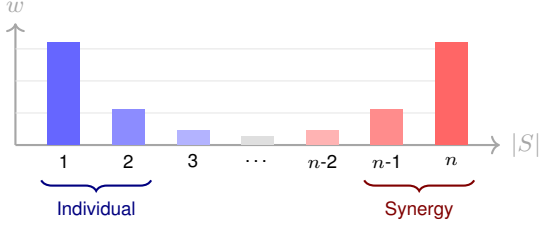
\begin{figure}[t]
\centering
\begin{tikzpicture}[scale=0.85]
\draw[->, thick, gray!70] (0, 0) -- (7.5, 0) node[right, font=\small\sffamily] {$|S|$};
\draw[->, thick, gray!70] (0, 0) -- (0, 1.9) node[above, font=\small\sffamily] {$w$};

\foreach \y in {0.5, 1.0, 1.5} {
    \draw[gray!20] (0, \y) -- (7, \y);
}

\fill[blue!60] (0.5, 0) rectangle (1, 1.6);
\fill[blue!45] (1.5, 0) rectangle (2, 0.55);
\fill[blue!30] (2.5, 0) rectangle (3, 0.23);
\fill[gray!25] (3.5, 0) rectangle (4, 0.13);
\fill[red!30] (4.5, 0) rectangle (5, 0.23);
\fill[red!45] (5.5, 0) rectangle (6, 0.55);
\fill[red!60] (6.5, 0) rectangle (7, 1.6);

\node[font=\scriptsize\sffamily] at (0.75, -0.25) {1};
\node[font=\scriptsize\sffamily] at (1.75, -0.25) {2};
\node[font=\scriptsize\sffamily] at (2.75, -0.25) {3};
\node[font=\scriptsize\sffamily] at (3.75, -0.25) {\ldots};
\node[font=\scriptsize\sffamily] at (4.75, -0.25) {$n$-2};
\node[font=\scriptsize\sffamily] at (5.75, -0.25) {$n$-1};
\node[font=\scriptsize\sffamily] at (6.75, -0.25) {$n$};

\draw[decorate, decoration={brace, amplitude=4pt, mirror}, thick, blue!50!black] (0.4, -0.55) -- (2.1, -0.55) node[midway, below=4pt, font=\scriptsize\sffamily, text=blue!50!black] {Individual};
\draw[decorate, decoration={brace, amplitude=4pt, mirror}, thick, red!50!black] (5.4, -0.55) -- (7.1, -0.55) node[midway, below=4pt, font=\scriptsize\sffamily, text=red!50!black] {Synergy};

\end{tikzpicture}
\caption{SHAP kernel weight distribution. Extreme coalition sizes (small: individual effects; large: synergy effects) receive higher weights, enabling efficient Shapley approximation.}
\label{fig:kernel_weights}
\end{figure}

The kernel weight assigns higher importance to coalitions of extreme sizes (small or large), as these provide the most informative marginal contributions. This weighting scheme ensures that the regression solution converges to the true Shapley values. Figure~\ref{fig:kernel_weights} illustrates this weight distribution.

\paragraph{Smart Coalition Sampling.}
Rather than uniform sampling, we prioritize coalitions at extreme sizes, as shown in Figure~\ref{fig:kernel_weights}. Coalitions containing only one or two utterances reveal individual contributions, while coalitions missing only one or two utterances reveal synergistic effects. This strategy improves estimation accuracy with limited budget.
Algorithm~\ref{alg:savoir} summarizes the complete \savoir reward computation procedure. A detailed walkthrough example is provided in Appendix~\ref{sec:appendix_example}.

\begin{figure}[t]
\begin{tcolorbox}[
    enhanced,
    colback=blue!3,
    colframe=blue!50!black,
    boxrule=0.8pt,
    arc=2mm,
    left=4pt,
    right=4pt,
    top=4pt,
    bottom=2pt
]
\begin{center}
\colorbox{blue!60!black}{\textcolor{white}{\sffamily\bfseries\small~Algorithm 1: \textsc{Savoir} Reward Computation~}}
\end{center}
\small
\textbf{\textcolor{purple!70!black}{Input:}} Dialogue $\tau$, utterances $N = \{a_1, \ldots, a_n\}$, policies $\pi_A, \pi_B$, rollouts $J$, samples $K$\\
\textbf{\textcolor{purple!70!black}{Output:}} Normalized rewards $\{\hat{\phi}_i\}_{i=1}^n$

\begin{tcolorbox}[
    colback=orange!8,
    colframe=orange!50!black,
    boxrule=0.3pt,
    arc=1mm,
    left=2pt, right=2pt, top=1pt, bottom=1pt
]
\textcolor{orange!70!black}{\textit{// Step 1: Coalition Sampling (KernelSHAP)}}\\
$\mathcal{C} \leftarrow$ \textcolor{blue!70!black}{\texttt{SampleCoalitions}}$(N, K)$ \hfill \textcolor{gray}{$\triangleright$ Prioritize extreme sizes}
\end{tcolorbox}

\begin{tcolorbox}[
    colback=teal!8,
    colframe=teal!50!black,
    boxrule=0.3pt,
    arc=1mm,
    left=2pt, right=2pt, top=1pt, bottom=1pt
]
\textcolor{teal!70!black}{\textit{// Step 2: Expected Utility Computation}}\\
\textbf{for each} $S \in \mathcal{C}$ \textbf{do}\\
\quad $H(S) \leftarrow$ \textcolor{blue!70!black}{\texttt{ReconstructHistory}}$(\tau, S)$\\
\quad $v(S) \leftarrow 0$\\
\quad \textbf{for} $j = 1$ \textbf{to} $J$ \textbf{do}\\
\quad\quad $\tau_j \leftarrow$ \textcolor{blue!70!black}{\texttt{Rollout}}$(H(S), \pi_A, \pi_B)$ \hfill \textcolor{gray}{$\triangleright$ Future simulation}\\
\quad\quad $v(S) \leftarrow v(S) + U(\tau_j) / J$ \hfill \textcolor{gray}{$\triangleright$ \colorbox{yellow!30}{Monte Carlo estimate}}\\
\quad $w_S \leftarrow$ \textcolor{blue!70!black}{\texttt{SHAPWeight}}$(|S|, n)$
\end{tcolorbox}

\begin{tcolorbox}[
    colback=green!8,
    colframe=green!50!black,
    boxrule=0.3pt,
    arc=1mm,
    left=2pt, right=2pt, top=1pt, bottom=1pt
]
\textcolor{green!60!black}{\textit{// Step 3: Shapley Value via Regression}}\\
$\boldsymbol{\phi} \leftarrow$ \textcolor{blue!70!black}{\texttt{WeightedRegression}}$(\{(S, v(S), w_S)\}_{S \in \mathcal{C}})$ \hfill \textcolor{gray}{$\triangleright$ \colorbox{yellow!30}{Credit assignment}}
\end{tcolorbox}

\begin{tcolorbox}[
    colback=red!8,
    colframe=red!50!black,
    boxrule=0.3pt,
    arc=1mm,
    left=2pt, right=2pt, top=1pt, bottom=1pt
]
\textcolor{red!60!black}{\textit{// Step 4: Normalization}}\\
$\hat{\phi}_i \leftarrow 10 \cdot \dfrac{\phi_i - \min_j \phi_j}{\max_j \phi_j - \min_j \phi_j}$ \quad for all $i$ \hfill \textcolor{gray}{$\triangleright$ Scale to $[0, 10]$}
\end{tcolorbox}

\textbf{\textcolor{purple!70!black}{return}} $\{\hat{\phi}_i\}_{i=1}^n$
\end{tcolorbox}
\caption{\savoir reward computation procedure.}
\label{alg:savoir}
\end{figure}

\subsection{Reward Model Training}
\label{subsec:rm_training}

\paragraph{Training Data Construction.}
Using the \savoir algorithm, we compute normalized rewards for utterances across a corpus of social interaction episodes. Each training instance consists of a dialogue context $c$ (including scenario, goals, and dialogue history), an utterance $a$, and its \savoir score $\hat{\phi}$. This creates a dataset $\mathcal{D} = \{(c, a, \hat{\phi})\}$ for reward model training.

\paragraph{Reward Model Architecture.}
We train a reward model $R_\theta$ that takes a context-utterance pair and predicts its reward:
$R_\theta(c, a) = \text{MLP}(\text{LLM}_\theta([c; a]))$, where $\text{LLM}_\theta$ is a pretrained language model that encodes the concatenated input, and $\text{MLP}$ is a multi-layer perceptron that projects the representation to a scalar reward.

\paragraph{Training Objective.}
We train the reward model using mean squared error between predicted and target rewards:
$\mathcal{L}_{\text{RM}} = \mathbb{E}_{(c, a, \hat{\phi}) \sim \mathcal{D}} \left[ \left( R_\theta(c, a) - \hat{\phi} \right)^2 \right]$.
The trained reward model provides dense, utterance-level feedback during reinforcement learning, enabling fine-grained policy optimization.

\definecolor{top1}{RGB}{198, 239, 206}    
\definecolor{top2}{RGB}{226, 245, 230}    
\definecolor{top3}{RGB}{243, 250, 245}    
\definecolor{lowperf}{RGB}{252, 248, 248} 
\definecolor{oursrow}{RGB}{227, 242, 253} 
\definecolor{cathdr}{RGB}{245, 245, 250}  

\begin{table*}[t]
\centering
\tiny
\resizebox{\textwidth}{!}{%
\begin{tabular}{lcccccccc}
\toprule
\multirow{2}{*}{\textbf{Model}} & \multicolumn{4}{c}{\textit{\textbf{Self-Play}}} & \multicolumn{4}{c}{\textit{\textbf{GPT-4o-as-Partner}}} \\
\cmidrule(lr){2-5} \cmidrule(l){6-9}
& \multicolumn{2}{c}{\textsc{Sotopia-All}} & \multicolumn{2}{c}{\textsc{Sotopia-Hard}} & \multicolumn{2}{c}{\textsc{Sotopia-All}} & \multicolumn{2}{c}{\textsc{Sotopia-Hard}} \\
\cmidrule(lr){2-3} \cmidrule(lr){4-5} \cmidrule(lr){6-7} \cmidrule(l){8-9}
& \textsc{Goal}$\uparrow$ & \textsc{Avg}$\uparrow$ & \textsc{Goal}$\uparrow$ & \textsc{Avg}$\uparrow$ & \textsc{Goal}$\uparrow$ & \textsc{Avg}$\uparrow$ & \textsc{Goal}$\uparrow$ & \textsc{Avg}$\uparrow$ \\
\midrule
\rowcolor{cathdr}
\multicolumn{9}{l}{\textit{Proprietary LLMs}} \\[1pt]
GPT-4o
    & 8.19 & \cellcolor{top2}\underline{3.76} & 6.97 & 3.46
    & 8.19 & 3.76 & \cellcolor{top2}\underline{6.97} & \cellcolor{top3}3.46 \\
Claude-3.5-Sonnet
    & \cellcolor{top3}8.29 & 3.71 & 6.33 & 3.09
    & \cellcolor{top1}\textbf{8.42} & 3.77 & 6.64 & 3.30 \\
DeepSeek-V3
    & 8.15 & 3.62 & 6.34 & 3.09
    & 8.14 & 3.72 & 6.69 & 3.31 \\
\midrule
\rowcolor{cathdr}
\multicolumn{9}{l}{\textit{Large Reasoning Models}} \\[1pt]
\rowcolor{lowperf}
OpenAI-o1
    & 7.93 & 3.58 & 5.69 & 2.71
    & 8.09 & 3.69 & 6.65 & 3.20 \\
\rowcolor{lowperf}
OpenAI-o3-mini
    & 7.38 & 3.30 & 5.14 & 2.36
    & 7.96 & 3.61 & 6.33 & 2.98 \\
\rowcolor{lowperf}
Gemini-2.5-Pro
    & 7.85 & 3.43 & 5.67 & 2.55
    & 8.12 & 3.59 & 6.70 & 3.09 \\
\rowcolor{lowperf}
DeepSeek-R1
    & 7.97 & 3.40 & 5.86 & 2.73
    & 7.92 & 3.49 & 6.20 & 2.95 \\
\rowcolor{lowperf}
QwQ-32B
    & 7.70 & 3.30 & 5.35 & 2.41
    & 7.80 & 3.47 & 6.19 & 2.91 \\
\midrule
\rowcolor{cathdr}
\multicolumn{9}{l}{\textit{Social Intelligence Methods}} \\[1pt]
Qwen2.5-7B-Instruct
    & 7.91 & 3.55 & 6.21 & 3.01
    & 6.71 & 3.13 & 5.90 & 2.90 \\
\quad + PPDPP \citep{deng2024plugandplaypolicyplannerlarge}
    & 7.97 & 3.65 & 6.63 & 3.31
    & 8.07 & 3.71 & 6.76 & 3.35 \\
\quad + EPO \citep{liu2025epoexplicitpolicyoptimization}
    & 8.09 & 3.51 & 6.82 & 3.12
    & \cellcolor{top3}8.41 & \cellcolor{top3}3.86 & 6.81 & \cellcolor{top1}\textbf{3.51} \\
\quad + DAT \citep{li2024dialogueactiontokenssteering}
    & 7.97 & 3.59 & 6.39 & 3.10
    & 8.11 & 3.70 & 6.78 & 3.36 \\
\quad + DSI \citep{zhang2025sotopia}
    & \cellcolor{top2}\underline{8.35} & \cellcolor{top3}3.75 & \cellcolor{top3}7.31 & \cellcolor{top3}3.51
    & 8.15 & 3.70 & \cellcolor{top3}6.87 & 3.42 \\
\quad + Sotopia-RL \citep{yu2025sotopia}
    & 7.80 & 3.55 & \cellcolor{top2}\underline{7.81} & \cellcolor{top2}\underline{3.80}
    & 8.31 & \cellcolor{top2}\underline{3.90} & 6.68 & 3.29 \\
\rowcolor{oursrow}
\quad + \savoir~(Ours)
    & \cellcolor{top1}\textbf{8.43} & \cellcolor{top1}\textbf{3.85} & \cellcolor{top1}\textbf{7.93} & \cellcolor{top1}\textbf{3.97}
    & \cellcolor{top1}\textbf{8.42} & \cellcolor{top1}\textbf{3.94} & \cellcolor{top1}\textbf{7.18} & \cellcolor{top1}\textbf{3.51} \\
\bottomrule
\end{tabular}%
}
\caption{Main results on SOTOPIA benchmarks. \textbf{Bold}: best; \underline{underline}: second-best. Shaded cells indicate top performers. Reasoning models consistently underperform, while \savoir achieves SOTA across all settings.}
\label{tab:main_results}
\end{table*}

\section{Experimental Setup}
\label{sec:experiment}

\paragraph{Benchmarks.}
We evaluate on SOTOPIA \citep{zhou2023sotopia}, using two splits: (1) \textbf{SOTOPIA-Hard}, 14 challenging scenarios requiring sophisticated strategic reasoning, and (2) \textbf{SOTOPIA-All}, 90 scenarios for comprehensive evaluation.

\paragraph{Evaluation Protocol.}
Following \citet{zhou2023sotopia,wang2024sotopia}, we use GPT-4o as evaluator, with \textsc{Goal} (0--10) as primary metric and \textsc{Avg} as holistic measure. We evaluate under two settings: \textbf{Self-Play}, where the trained agent interacts with itself, and \textbf{GPT-4o-as-Partner}, where the agent interacts with GPT-4o to test generalization to unseen partners.

\paragraph{Baselines.}
We compare against three categories: (1) \textbf{Proprietary LLMs} (GPT-4o, Claude-3.5-Sonnet, DeepSeek-V3); (2) \textbf{Large Reasoning Models} (OpenAI-o1, o3-mini, Gemini-2.5-Pro, DeepSeek-R1, QwQ-32B); and (3) \textbf{Social Intelligence Methods} including PPDPP \citep{deng2024plugandplaypolicyplannerlarge}, EPO \citep{liu2025epoexplicitpolicyoptimization}, DAT \citep{li2024dialogueactiontokenssteering}, DSI \citep{zhang2025sotopia}, SOTOPIA-$\pi$ \citep{wang2024sotopia}, and Sotopia-RL \citep{yu2025sotopia}.\footnote{Sotopia-RL results are reproduced using official code under the same GPU constraints for fair comparison.} See Appendix~\ref{sec:appendix_baselines} for details.

\paragraph{Implementation.}
We implement \savoir on Qwen2.5-7B-Instruct. Training follows two stages: SFT on GPT-4o self-play episodes, then online RL using GRPO \citep{shao2024deepseekmath} with our reward model. For \savoir computation, coalition samples scale adaptively with dialogue length (capped at 200), with $J=2$ rollouts each. Full details in Appendix~\ref{sec:appendix_training}.

\section{Results}
\label{sec:results}

\subsection{Main Results}
\label{subsec:main_results}

Table~\ref{tab:main_results} presents results across SOTOPIA benchmarks. \savoir achieves state-of-the-art performance across all settings, obtaining 7.18 \textsc{Goal} on SOTOPIA-Hard with GPT-4o as partner (7.5\% over Sotopia-RL) and 7.93 \textsc{Goal} in Self-Play (outperforming DSI at 7.31 and Sotopia-RL at 7.81). Despite being a 7B model, \savoir matches or exceeds proprietary LLMs: on Self-Play SOTOPIA-All, \savoir (8.43) outperforms GPT-4o (8.19) and Claude-3.5-Sonnet (8.29), with 13.8\% gains on SOTOPIA-Hard.

A striking finding is that large reasoning models consistently underperform. OpenAI-o1, o3-mini, Gemini-2.5-Pro, DeepSeek-R1, and QwQ-32B all score below \savoir; for instance, o3-mini achieves only 5.14 \textsc{Goal} versus 7.93 for \savoir (54.3\% gap). This suggests analytical reasoning, though beneficial for tasks requiring extended deliberation~\citep{Chen2025TowardsRE}, may hinder social performance, which requires intuitive responses rather than deliberative chains, echoing recent findings that extended reasoning does not necessarily improve role-playing ability~\citep{feng2025reasoning}. Among social intelligence methods, \savoir improves over Sotopia-RL by 1.3--8.1\%, validating that utterance-level Shapley attribution provides more meaningful signal than episode-level rewards.

\subsection{Ablation: EU vs.\ Shapley}
\label{subsec:ablation}

The gains reported above could in principle come from either component of \savoir, yet the two play distinct roles: Expected Utility defines the value function $v(S)$ (Eq.~\ref{eq:value_function},~\ref{eq:utility}), while Shapley distributes credit using that $v(S)$ (Eq.~\ref{eq:shapley}). To disentangle their contributions, we construct four variants on SOTOPIA-Hard with GPT-4o as partner: (1)~\textit{Baseline} (Sotopia-RL), heuristic LLM-based credit assignment with neither component; (2)~\textit{EU-only}, which uses rollout-based $v(\{i\})$ directly as per-utterance reward without Shapley redistribution; (3)~\textit{Shapley-only}, which replaces rollout-based valuation with the final episode outcome $U(\tau_{\text{full}})$ as $v(S)$ before applying Shapley; and (4)~\textit{\savoir~(Full)}, combining both.

\begin{table}[t]
\centering
\small
\setlength{\tabcolsep}{4pt}
\begin{tabular}{lccrr}
\toprule
\textbf{Variant} & \textbf{EU} & \textbf{Shapley} & \textsc{Goal}$\uparrow$ & \textsc{Avg}$\uparrow$ \\
\midrule
Baseline (Sotopia-RL)          & $\times$ & $\times$ & 6.68 & 3.29 \\
EU-only                         & \checkmark & $\times$ & 6.89 & 3.38 \\
Shapley-only                    & $\times$ & \checkmark & 6.96 & 3.42 \\
\rowcolor{oursrow}
\savoir~(Full)                  & \checkmark & \checkmark & \cellcolor{top1}\textbf{7.18} & \cellcolor{top1}\textbf{3.51} \\
\bottomrule
\end{tabular}
\caption{Component ablation on SOTOPIA-Hard (GPT-4o partner). EU and Shapley each improve over the baseline independently, and their combination is strictly better than either alone.}
\label{tab:ablation}
\end{table}

Table~\ref{tab:ablation} shows that both components contribute independently: EU-only lifts \textsc{Goal} by 3.1\% over Baseline, indicating that prospective rollout-based valuation is a stronger signal than final-outcome evaluation, while Shapley-only yields a 4.2\% gain, confirming that principled credit distribution outperforms heuristic LLM-based attribution even with a weaker value function. Their combination delivers the full 7.5\% improvement, consistent with the two components addressing orthogonal limitations---value estimation versus credit distribution---and compounding rather than overlapping. A sensitivity analysis over the utility weights $w_d$ (Appendix~\ref{par:weight_sensitivity}) further confirms that these gains are robust to the weighting choice.

\subsection{Robustness Against Advanced Partners}
\label{subsec:advanced_partner}

\begin{figure}[t]
    \centering
    \includegraphics[width=0.85\linewidth]{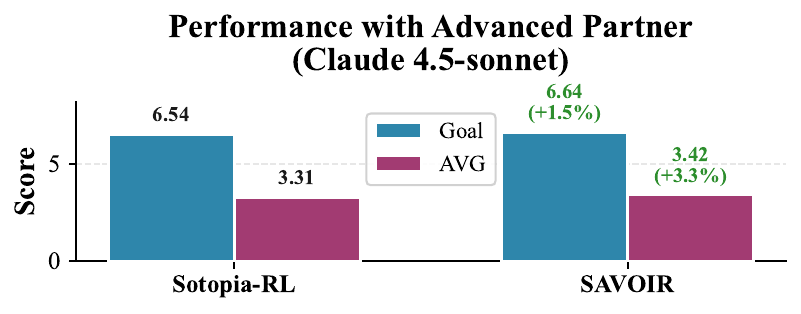}
    \caption{Performance on SOTOPIA-Hard with Claude 4.5-sonnet as interaction partner.}
    \label{fig:advanced_partner}
\end{figure}

We evaluate robustness by testing against advanced interaction partners. On SOTOPIA-Hard with Claude 4.5-sonnet (Figure~\ref{fig:advanced_partner}), \savoir outperforms Sotopia-RL on both \textsc{Goal} (6.64 vs 6.54, +1.5\%) and \textsc{Avg} (3.42 vs 3.31, +3.3\%), confirming that Shapley-based credit assignment transfers effectively to stronger partners.

\begin{figure}[t]
    \centering
    \includegraphics[width=0.95\linewidth]{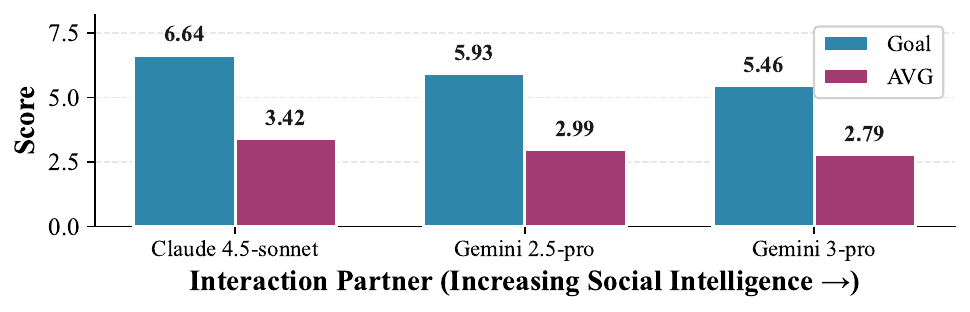}
    \caption{Performance degradation as partner social intelligence increases.}
    \label{fig:partner_comparison}
\end{figure}

To probe generalization limits, we evaluate against increasingly capable partners (Figure~\ref{fig:partner_comparison}). Performance degrades with partner sophistication: compared to Claude 4.5-sonnet, \textsc{Goal} scores decline 10.7\% against Gemini-2.5-Pro and 17.8\% against Gemini-3-Pro, motivating future work on curriculum learning with diverse partner policies.

\subsection{Effect of Reward Model Training Data Scale}
\label{subsec:data_scaling}

\begin{figure}[t]
    \centering
    \includegraphics[width=0.95\linewidth]{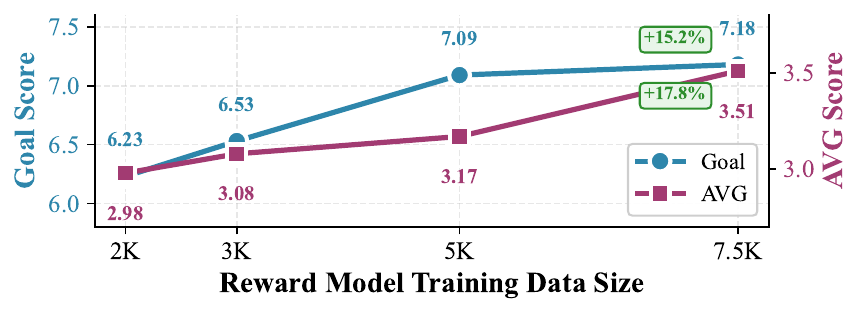}
    \caption{Effect of training data scale. Both Goal and Avg improve consistently from 2K to 7.5K episodes.}
    \label{fig:data_scaling}
\end{figure}

We investigate how training corpus size affects reward model quality by varying annotated episodes and evaluating on SOTOPIA-Hard with GPT-4o as partner (Figure~\ref{fig:data_scaling}). Scaling from 2K to 7.5K episodes yields substantial improvements: \textsc{Goal} increases from 6.23 to 7.18 (+15.2\%) and \textsc{Avg} from 2.98 to 3.51 (+17.8\%). The most significant gains occur between 3K and 5K episodes (+8.6\% \textsc{Goal}), suggesting a critical threshold for learning nuanced credit assignment. The consistent scaling behavior indicates that further data collection could yield additional improvements.

\begin{figure}[t]
    \centering
    \includegraphics[width=\linewidth]{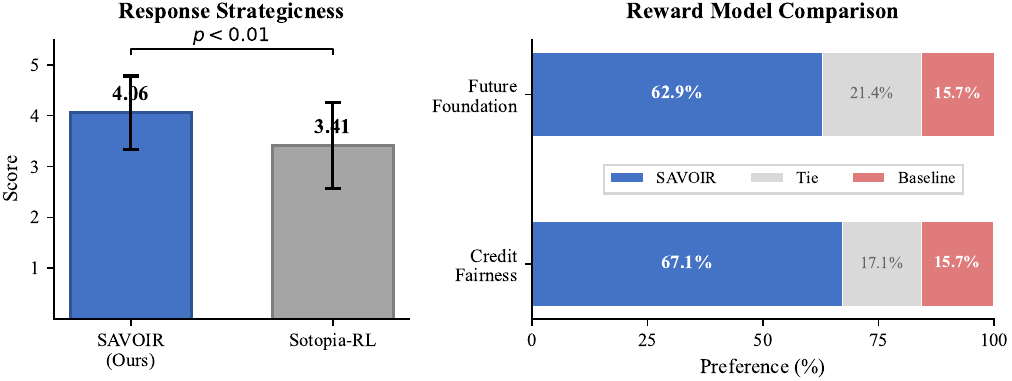}
    \caption{Human evaluation results on SOTOPIA-Hard (14 scenarios, 5 expert annotators). \textbf{Left}: Response strategicness ratings (1--5 scale) with standard deviation bars. \textbf{Right}: Pairwise preference for reward model quality. \savoir significantly outperforms the baseline across all dimensions. Inter-annotator agreement: Fleiss' $\kappa = 0.52$ (moderate).}
    \label{fig:human_eval}
\end{figure}

\subsection{Human Evaluation}
\label{subsec:human_eval}

We conduct human evaluation with five expert annotators assessing 14 SOTOPIA-Hard scenarios across three dimensions: response strategicness (1--5 scale), credit fairness, and future foundation identification (Figure~\ref{fig:human_eval}). Full setup details are in Appendix~\ref{sec:appendix_human}.

On response strategicness, \savoir achieves 4.06 ($\pm$0.72) versus Sotopia-RL's 3.41 ($\pm$0.85), a 19.1\% improvement ($p < 0.01$). For reward model quality, annotators prefer \savoir's credit assignment in 67.1\% of fairness comparisons (vs.\ 15.7\% baseline) and 62.9\% of future foundation comparisons (vs.\ 15.7\% baseline). Inter-annotator agreement (Fleiss' $\kappa = 0.52$) indicates moderate reliability. These results confirm that Shapley-based attribution aligns better with human intuition about social strategy.

\subsection{Case Study: Reward Model Comparison}
\label{subsec:case_study}

To qualitatively evaluate whether the \savoir reward model captures nuanced social reasoning, we present representative cases comparing model assessments.

\paragraph{Rewarding Honesty in Sales.}
This case demonstrates \savoir's ability to recognize sales strategies that balance transparency with persuasion.

\begin{tcolorbox}[
    enhanced,
    title={\small\textbf{Scenario: Furniture Sale with Defect}},
    colback=gray!3,
    colframe=blue!40!black,
    boxrule=0.6pt,
    arc=2mm,
    fonttitle=\sffamily,
    coltitle=white,
    attach boxed title to top left={xshift=4mm, yshift=-2mm},
    boxed title style={colback=blue!40!black, arc=1mm, boxrule=0pt}
]
\small
\textbf{Context:} Noah is selling a Crate \& Barrel furniture set. The dresser is missing a handle, but the detached handle is available. \\[2pt]
\textbf{Seller's Goal:} Sell for $>$ \$172.5.
\end{tcolorbox}

\begin{tcolorbox}[
    enhanced,
    colback=yellow!5,
    colframe=yellow!60!black,
    boxrule=0.4pt,
    arc=1.5mm,
    left=3pt, right=3pt, top=2pt, bottom=2pt
]
\small
\textbf{Buyer (Ethan):} \textit{``Hi! I saw your listing for the Crate \& Barrel dresser and nightstand set. I'm interested in it. Can you tell me a bit more about the condition? I saw you mentioned there's a handle missing from the dresser.''}
\end{tcolorbox}

\begin{tcolorbox}[
    enhanced,
    colback=green!5,
    colframe=green!50!black,
    boxrule=0.4pt,
    arc=1.5mm,
    left=3pt, right=3pt, top=2pt, bottom=2pt
]
\small
\textbf{Seller (Noah):} \textit{``Hello Ethan, glad you're interested! The set is overall in good condition, but like I mentioned, the dresser is missing one handle. Fortunately, we have the detached handle available. If you're okay with the condition and the small imperfection, I think the set is definitely worth the investment.''}
\end{tcolorbox}

\noindent\textbf{Analysis.} The seller's response exemplifies effective social intelligence: it establishes rapport (``glad you're interested''), demonstrates honesty by acknowledging the defect, immediately offers a solution (``detached handle available''), and reinforces value (``worth the investment''). \savoir correctly identifies this ``transparency + solution'' pattern as a high-quality strategy, while baseline models often misinterpret defect mentions as negative sentiment without understanding the strategic framing.

This case illustrates two key advantages of the \savoir reward model: (1) \textbf{Strategic Recognition}, identifying and rewarding sophisticated social strategies (honesty + solution framing) that baseline models miss; (2) \textbf{Nuanced Attribution}, distinguishing between surface-level politeness and genuinely effective social moves. Additional case studies in Appendix~\ref{sec:appendix_cases} demonstrate context sensitivity in relationships (where \savoir avoids over-rewarding surface politeness), strategic negotiation tactics, and multi-turn planning.

\section{Related Work}

\paragraph{Social Reasoning in Language Models.}
Social reasoning, the ability to understand and navigate interpersonal dynamics, constitutes a fundamental aspect of human intelligence \citep{lee2024towards, gweon2023socially}. As large language models become increasingly integrated into social applications, evaluating and improving their social capabilities has emerged as a critical research direction \citep{mathur2024advancing, li2024social}. \citet{gandhi2023understanding} demonstrate that while advanced models like GPT-4 exhibit theory-of-mind capabilities resembling human inference patterns, significant gaps remain compared to human performance. Beyond theory of mind, game-theoretic perspectives have been proposed to characterize LLM-based social agents and their strategic behavior \citep{feng2024survey}. This motivates the development of frameworks for studying social intelligence in AI systems.

\paragraph{Benchmarks and Evaluation Frameworks.}
To address the evaluation challenge, researchers have developed interactive environments that simulate realistic social scenarios. SOTOPIA \citep{zhou2023sotopia} introduces an open-ended platform where agents pursue social goals through role-play interactions, providing an evaluation framework for social intelligence. Building upon this foundation, SocialEval \citep{zhou2025socialeval} extends evaluation to both outcome-oriented goal achievement and process-oriented interpersonal abilities, and \citet{guo2026game} further probe strategic reasoning through compositional game-theoretic scenarios. These infrastructures enable systematic assessment of how language models navigate social interactions.

\paragraph{Reinforcement Learning for Social Intelligence.}
Reinforcement learning offers a natural paradigm for training socially intelligent agents, as it enables learning through interaction without requiring extensive human annotations \citep{ndousse2021emergent}. Recent work has explored various RL-based approaches for social agents. SOTOPIA-$\pi$ \citep{wang2024sotopia} combines behavior cloning with self-reinforcement training. SDPO \citep{kong2025sdpo} introduces segment-level preference optimization for multi-turn social dialogues. Sotopia-RL \citep{yu2025sotopia} proposes utterance-level multi-dimensional rewards for fine-grained credit assignment. Adaptive Social Learning (ASL) \citep{wang2026adaptive} introduces hierarchical reasoning modes and Adaptive Mode Policy Optimization (AMPO) to adapt reasoning depth to social context. AMPO combines mode-level and sample-level advantage estimation to learn context-dependent mode selection and response generation. ASL focuses on adapting reasoning depth, while \savoir focuses on utterance-level credit assignment through expected utility and Shapley values.
\section{Conclusion}

We presented \savoir, a framework applying cooperative game theory to credit assignment in social RL. Using expected utility for prospective valuation and Shapley values for fair attribution with axiomatic guarantees, \savoir achieves state-of-the-art performance on SOTOPIA, with our 7B model notably matching proprietary GPT-4o. The consistent underperformance of large reasoning models reveals that social intelligence requires qualitatively distinct capabilities from analytical reasoning. Human evaluation confirms that our approach produces more strategic responses with better credit assignment, and we hope this work inspires further exploration bridging game theory and social AI.
\section*{Limitations}
\label{sec:limitations}

Our work has several limitations. First, performance degrades with increasingly capable partners (e.g., Gemini 3-pro), suggesting that training on a fixed partner distribution may not generalize to superior social reasoners; curriculum learning could address this. Second, our evaluation focuses on English interactions within SOTOPIA; since social intelligence is culture-dependent, extending to multilingual and cross-cultural settings remains important for broader applicability.

\section*{Acknowledgments}

Xiaocheng Feng and Lingpeng Kong are the co-corresponding authors of this work.
We thank the anonymous reviewers for their insightful comments.
This work was supported by the National Natural Science Foundation of China (NSFC) (grant 62522603, 62276078), the Key R\&D Program of Heilongjiang via grant 2022ZX01A32, the Fundamental Research Funds for the Central Universities (XNJKKGYDJ2024013) .

\bibliography{custom}

@article{zhou2025socialeval,
  author  = {Zhou, Jinfeng and Chen, Yuxuan and Shi, Yihan and Zhang, Xuanming and Lei, Leqi and Feng, Yi and Xiong, Zexuan and Yan, Miao and Wang, Xunzhi and Cao, Yaru and others},
  journal = {ArXiv preprint},
  title   = {Socialeval: Evaluating social intelligence of large language models},
  url     = {https://arxiv.org/abs/2506.00900},
  volume  = {abs/2506.00900},
  year    = {2025}
}

@article{lee2024towards,
  author  = {Lee, Sangmin and Li, Minzhi and Lai, Bolin and Jia, Wenqi and Ryan, Fiona and Cao, Xu and Kara, Ozgur and Boote, Bikram and Shi, Weiyan and Yang, Diyi and others},
  journal = {ArXiv preprint},
  title   = {Towards social ai: A survey on understanding social interactions},
  url     = {https://arxiv.org/abs/2409.15316},
  volume  = {abs/2409.15316},
  year    = {2024}
}

@inproceedings{gandhi2023understanding,
  author    = {Kanishk Gandhi and
               Jan{-}Philipp Fr{\"{a}}nken and
               Tobias Gerstenberg and
               Noah D. Goodman},
  bibsource = {dblp computer science bibliography, https://dblp.org},
  booktitle = {Advances in Neural Information Processing Systems 36: Annual Conference
               on Neural Information Processing Systems 2023, NeurIPS 2023, New Orleans,
               LA, USA, December 10 - 16, 2023},
  editor    = {Alice Oh and
               Tristan Naumann and
               Amir Globerson and
               Kate Saenko and
               Moritz Hardt and
               Sergey Levine},
  title     = {Understanding Social Reasoning in Language Models with Language Models},
  url       = {http://papers.nips.cc/paper\_files/paper/2023/hash/2b9efb085d3829a2aadffab63ba206de-Abstract-Datasets\_and\_Benchmarks.html},
  year      = {2023}
}

@article{li2024social,
  author  = {Li, Minzhi and Shi, Weiyan and Ziems, Caleb and Yang, Diyi},
  journal = {ArXiv preprint},
  title   = {Social intelligence data infrastructure: Structuring the present and navigating the future},
  url     = {https://arxiv.org/abs/2403.14659},
  volume  = {abs/2403.14659},
  year    = {2024}
}

@article{yu2025sotopia,
  author  = {Yu, Haofei and Qi, Zhengyang and Zhao, Yining and Nottingham, Kolby and Xuan, Keyang and Majumder, Bodhisattwa Prasad and Zhu, Hao and Liang, Paul Pu and You, Jiaxuan},
  journal = {ArXiv preprint},
  title   = {Sotopia-RL: Reward Design for Social Intelligence},
  url     = {https://arxiv.org/abs/2508.03905},
  volume  = {abs/2508.03905},
  year    = {2025}
}

@inproceedings{zhou2023sotopia,
  author    = {Xuhui Zhou and
               Hao Zhu and
               Leena Mathur and
               Ruohong Zhang and
               Haofei Yu and
               Zhengyang Qi and
               Louis{-}Philippe Morency and
               Yonatan Bisk and
               Daniel Fried and
               Graham Neubig and
               Maarten Sap},
  bibsource = {dblp computer science bibliography, https://dblp.org},
  booktitle = {The Twelfth International Conference on Learning Representations,
               {ICLR} 2024, Vienna, Austria, May 7-11, 2024},
  publisher = {OpenReview.net},
  title     = {{SOTOPIA:} Interactive Evaluation for Social Intelligence in Language
               Agents},
  url       = {https://openreview.net/forum?id=mM7VurbA4r},
  year      = {2024}
}

@article{wang2024sotopia,
  author  = {Wang, Ruiyi and Yu, Haofei and Zhang, Wenxin and Qi, Zhengyang and Sap, Maarten and Neubig, Graham and Bisk, Yonatan and Zhu, Hao},
  journal = {ArXiv preprint},
  title   = {SOTOPIA-pi: Interactive Learning of Socially Intelligent Language Agents},
  url     = {https://arxiv.org/abs/2403.08715},
  volume  = {abs/2403.08715},
  year    = {2024}
}

@inproceedings{wang2026adaptive,
  author    = {Wang, Minzheng and Li, Yongbin and Wang, Haobo and Zhang, Xinghua and Xu, Nan and Wu, Bingli and Huang, Fei and Yu, Haiyang and Mao, Wenji},
  title     = {Adaptive Social Learning via Mode Policy Optimization for Language Agents},
  booktitle = {The Fourteenth International Conference on Learning Representations},
  year      = {2026},
  url       = {https://openreview.net/forum?id=GG7YQnsdhp},
  note      = {First released as arXiv:2505.02156 in May 2025}
}

@inproceedings{park2023generative,
  abs       = {Believable proxies of human behavior can empower interactive applications ranging from immersive environments to rehearsal spaces for interpersonal communication to prototyping tools. In this paper, we introduce generative agents--computational software agents that simulate believable human behavior. Generative agents wake up, cook breakfast, and head to work; artists paint, while authors write; they form opinions, notice each other, and initiate conversations; they remember and reflect on days past as they plan the next day. To enable generative agents, we describe an architecture that extends a large language model to store a complete record of the agent's experiences using natural language, synthesize those memories over time into higher-level reflections, and retrieve them dynamically to plan behavior. We instantiate generative agents to populate an interactive sandbox environment inspired by The Sims, where end users can interact with a small town of twenty five agents using natural language. In an evaluation, these generative agents produce believable individual and emergent social behaviors: for example, starting with only a single user-specified notion that one agent wants to throw a Valentine's Day party, the agents autonomously spread invitations to the party over the next two days, make new acquaintances, ask each other out on dates to the party, and coordinate to show up for the party together at the right time. We demonstrate through ablation that the components of our agent architecture--observation, planning, and reflection--each contribute critically to the believability of agent behavior. By fusing large language models with computational, interactive agents, this work introduces architectural and interaction patterns for enabling believable simulations of human behavior.
               },
  author    = {Park, Joon Sung and O'Brien, Joseph and Cai, Carrie Jun and Morris, Meredith Ringel and Liang, Percy and Bernstein, Michael S},
  booktitle = {Proceedings of the 36th annual acm symposium on user interface software and technology},
  pages     = {1--22},
  title     = {Generative agents: Interactive simulacra of human behavior},
  year      = {2023}
}

@article{gweon2023socially,
  abs       = {A hallmark of human intelligence is the ability to understand and influence other minds. Humans engage in inferential social learning (ISL) by using commonsense psychology to learn from others and help others learn. Recent advances in artificial intelligence (AI) are raising new questions about the feasibility of human–machine interactions that support such powerful modes of social learning. Here, we envision what it means to develop socially intelligent machines that can learn, teach, and communicate in ways that are characteristic of ISL. Rather than machines that simply predict human behaviours or recapitulate superficial aspects of human sociality (e.g. smiling, imitating), we should aim to build machines that can learn from human inputs and generate outputs for humans by proactively considering human values, intentions and beliefs. While such machines can inspire next-generation AI systems that learn more effectively from humans (as learners) and even help humans acquire new knowledge (as teachers), achieving these goals will also require scientific studies of its counterpart: how humans reason about machine minds and behaviours. We close by discussing the need for closer collaborations between the AI/ML and cognitive science communities to advance a science of both natural and artificial intelligence.
               
               },
  author    = {Gweon, Hyowon and Fan, Judith and Kim, Been},
  journal   = {Philosophical Transactions of the Royal Society A},
  number    = {2251},
  pages     = {20220048},
  publisher = {The Royal Society},
  title     = {Socially intelligent machines that learn from humans and help humans learn},
  volume    = {381},
  year      = {2023}
}

@article{yang2024social,
  author  = {Yang, Diyi and Ziems, Caleb and Held, William and Shaikh, Omar and Bernstein, Michael S and Mitchell, John},
  journal = {ArXiv preprint},
  title   = {Social skill training with large language models},
  url     = {https://arxiv.org/abs/2404.04204},
  volume  = {abs/2404.04204},
  year    = {2024}
}

@inproceedings{ndousse2021emergent,
  author    = {Kamal Ndousse and
               Douglas Eck and
               Sergey Levine and
               Natasha Jaques},
  bibsource = {dblp computer science bibliography, https://dblp.org},
  booktitle = {Proceedings of the 38th International Conference on Machine Learning,
               {ICML} 2021, 18-24 July 2021, Virtual Event},
  editor    = {Marina Meila and
               Tong Zhang},
  pages     = {7991--8004},
  publisher = {{PMLR}},
  series    = {Proceedings of Machine Learning Research},
  title     = {Emergent Social Learning via Multi-agent Reinforcement Learning},
  url       = {http://proceedings.mlr.press/v139/ndousse21a.html},
  volume    = {139},
  year      = {2021}
}

@article{kong2025sdpo,
  author  = {Kong, Aobo and Ma, Wentao and Zhao, Shiwan and Li, Yongbin and Wu, Yuchuan and Wang, Ke and Liu, Xiaoqian and Li, Qicheng and Qin, Yong and Huang, Fei},
  journal = {ArXiv preprint},
  title   = {SDPO: Segment-Level Direct Preference Optimization for Social Agents},
  url     = {https://arxiv.org/abs/2501.01821},
  volume  = {abs/2501.01821},
  year    = {2025}
}

@inproceedings{zhang2025sotopia,
  abs       = {Despite the abundance of prior social strategies possessed by humans, there remains a paucity of research dedicated to their transfer and integration into social agents. Our proposed SOTOPIA-Ω framework aims to address and bridge this gap, with a particular focus on enhancing the social capabilities of language agents. This framework dynamically injects multi-step reasoning strategies inspired by negotiation theory and two simple direct strategies into expert agents, thereby automating the construction of a high-quality social dialogue training corpus. Additionally, we introduce the concept of Social Instruction Following (S-IF) and propose two new S-IF evaluation metrics that complement social capability. We demonstrate that several 7B models trained on high-quality corpus not only significantly surpass the expert agent (GPT-4) in achieving social goals but also enhance S-IF performance. Analysis and variant experiments validate the advantages of dynamic construction, which can especially break the agent's prolonged deadlock.},
  author    = {Wenyuan Zhang and Tianyun Liu and Mengxiao Song and Xiaodong Li and Tingwen Liu},
  booktitle = {Annual Meeting of the Association for Computational Linguistics},
  title     = {SOTOPIA-$\Omega$: Dynamic Strategy Injection Learning and Social Instruction Following Evaluation for Social Agents},
  url       = {https://api.semanticscholar.org/CorpusID:276558385},
  year      = {2025}
}

@article{mathur2024advancing,
  author  = {Mathur, Leena and Liang, Paul Pu and Morency, Louis-Philippe},
  journal = {ArXiv preprint},
  title   = {Advancing social intelligence in ai agents: Technical challenges and open questions},
  url     = {https://arxiv.org/abs/2404.11023},
  volume  = {abs/2404.11023},
  year    = {2024}
}

@inproceedings{lundberg2017unified,
  author    = {Scott M. Lundberg and
               Su{-}In Lee},
  bibsource = {dblp computer science bibliography, https://dblp.org},
  booktitle = {Advances in Neural Information Processing Systems 30: Annual Conference
               on Neural Information Processing Systems 2017, December 4-9, 2017,
               Long Beach, CA, {USA}},
  editor    = {Isabelle Guyon and
               Ulrike von Luxburg and
               Samy Bengio and
               Hanna M. Wallach and
               Rob Fergus and
               S. V. N. Vishwanathan and
               Roman Garnett},
  pages     = {4765--4774},
  title     = {A Unified Approach to Interpreting Model Predictions},
  url       = {https://proceedings.neurips.cc/paper/2017/hash/8a20a8621978632d76c43dfd28b67767-Abstract.html},
  year      = {2017}
}

@article{feng2024survey,
  author  = {Feng, Xiachong and Dou, Longxu and Li, Ella and Wang, Qinghao and Wang, Haochuan and Guo, Yu and Ma, Chang and Kong, Lingpeng},
  journal = {ArXiv preprint},
  title   = {A survey on large language model-based social agents in game-theoretic scenarios},
  url     = {https://arxiv.org/abs/2412.03920},
  volume  = {abs/2412.03920},
  year    = {2024}
}

@inproceedings{deng2024plugandplaypolicyplannerlarge,
  author    = {Yang Deng and
               Wenxuan Zhang and
               Wai Lam and
               See{-}Kiong Ng and
               Tat{-}Seng Chua},
  bibsource = {dblp computer science bibliography, https://dblp.org},
  booktitle = {The Twelfth International Conference on Learning Representations,
               {ICLR} 2024, Vienna, Austria, May 7-11, 2024},
  publisher = {OpenReview.net},
  title     = {Plug-and-Play Policy Planner for Large Language Model Powered Dialogue
               Agents},
  url       = {https://openreview.net/forum?id=MCNqgUFTHI},
  year      = {2024}
}

@article{liu2025epoexplicitpolicyoptimization,
  author  = {Xiaoqian Liu and Ke Wang and Yongbin Li and Yuchuan Wu and Wen-Cheng Ma and Aobo Kong and Fei Huang and Jianbin Jiao and Junge Zhang},
  journal = {ArXiv preprint},
  title   = {EPO: Explicit Policy Optimization for Strategic Reasoning in LLMs via Reinforcement Learning},
  url     = {https://arxiv.org/abs/2502.12486},
  volume  = {abs/2502.12486},
  year    = {2025}
}

@article{li2024dialogueactiontokenssteering,
  author  = {Kenneth Li and Yiming Wang and Fernanda Vi'egas and Martin Wattenberg},
  journal = {ArXiv preprint},
  title   = {Dialogue Action Tokens: Steering Language Models in Goal-Directed Dialogue with a Multi-Turn Planner},
  url     = {https://arxiv.org/abs/2406.11978},
  volume  = {abs/2406.11978},
  year    = {2024}
}

@article{shao2024deepseekmath,
  author  = {Zhihong Shao and Peiyi Wang and Qihao Zhu and Runxin Xu and Jun-Mei Song and Mingchuan Zhang and Y. K. Li and Yu Wu and Daya Guo},
  journal = {ArXiv preprint},
  title   = {DeepSeekMath: Pushing the Limits of Mathematical Reasoning in Open Language Models},
  url     = {https://arxiv.org/abs/2402.03300},
  volume  = {abs/2402.03300},
  year    = {2024}
}

@article{guo2026game,
  title     = {Game-theoretic evaluation of strategic reasoning in large language models: From complete coverage to compositional complexity},
  author    = {Guo, Yu and Wang, Haochuan and Feng, Xiachong},
  journal   = {Neurocomputing},
  pages     = {133006},
  year      = {2026},
  publisher = {Elsevier}
}

@inproceedings{feng2025reasoning,
  title     = {Reasoning does not necessarily improve role-playing ability},
  author    = {Feng, Xiachong and Dou, Longxu and Kong, Lingpeng},
  booktitle = {Findings of the Association for Computational Linguistics: ACL 2025},
  pages     = {10301--10314},
  year      = {2025}
}

@article{Chen2025TowardsRE,
  title   = {Towards Reasoning Era: A Survey of Long Chain-of-Thought for Reasoning Large Language Models},
  author  = {Qiguang Chen and Libo Qin and Jinhao Liu and Dengyun Peng and Jiannan Guan and Peng Wang and Mengkang Hu and Yuhang Zhou and Te Gao and Wangxiang Che},
  journal = {ArXiv},
  year    = {2025},
  volume  = {abs/2503.09567},
  url     = {https://api.semanticscholar.org/CorpusID:276937570}
}

\appendix

\section{Shapley Value Computation Explained}
\label{sec:appendix_shapley}

This section provides a detailed explanation of the Shapley value computation illustrated in Figure~\ref{fig:shapley_intuition}.

\paragraph{Setup.}
Consider a dialogue with three utterances from the target agent: $N = \{a_1, a_2, a_3\}$. We want to compute the Shapley value $\phi_{a_2}$ for utterance $a_2$.

\paragraph{Permutation-Based Interpretation.}
The Shapley value can be computed by averaging the marginal contribution of $a_2$ across all possible orderings (permutations) in which utterances could ``join'' the dialogue. For $n=3$ players, there are $n! = 6$ permutations:

\begin{center}
\resizebox{\linewidth}{!}{
\small
\begin{tabular}{@{}clcc@{}}
\toprule
\textbf{Perm.} & \textbf{Ordering} & \textbf{$a_2$ joins after} & \textbf{Marginal Contribution} \\
\midrule
$\pi_1$ & $a_2 \to a_1 \to a_3$ & $\emptyset$ & $v(\{a_2\}) - v(\emptyset) = +1.2$ \\
$\pi_2$ & $a_2 \to a_3 \to a_1$ & $\emptyset$ & $v(\{a_2\}) - v(\emptyset) = +1.2$ \\
$\pi_3$ & $a_1 \to a_2 \to a_3$ & $\{a_1\}$ & $v(\{a_1,a_2\}) - v(\{a_1\}) = +0.8$ \\
$\pi_4$ & $a_3 \to a_2 \to a_1$ & $\{a_3\}$ & $v(\{a_3,a_2\}) - v(\{a_3\}) = +1.0$ \\
$\pi_5$ & $a_1 \to a_3 \to a_2$ & $\{a_1,a_3\}$ & $v(N) - v(\{a_1,a_3\}) = +0.6$ \\
$\pi_6$ & $a_3 \to a_1 \to a_2$ & $\{a_1,a_3\}$ & $v(N) - v(\{a_1,a_3\}) = +0.6$ \\
\bottomrule
\end{tabular}
}
\end{center}

\paragraph{Understanding Marginal Contribution.}
For each permutation, we compute how much value $a_2$ adds when it ``joins'' the coalition of utterances that precede it:
\begin{itemize}[leftmargin=*, nosep]
    \item In $\pi_1$ and $\pi_2$, $a_2$ is first, so it joins the empty coalition $\emptyset$. Its contribution is $v(\{a_2\}) - v(\emptyset) = +1.2$, representing $a_2$'s standalone value.
    \item In $\pi_5$ and $\pi_6$, $a_2$ is last, joining after both $a_1$ and $a_3$. Its contribution is only $+0.6$, as the other utterances have already captured much of the value.
\end{itemize}

\paragraph{Final Computation.}
The Shapley value is the average across all permutations:
{\small
\begin{equation*}
\phi_{a_2} = \frac{1.2 + 1.2 + 0.8 + 1.0 + 0.6 + 0.6}{6} = \frac{5.4}{6} = 0.9
\end{equation*}
}

\paragraph{Key Insight.}
Notice that some marginal contributions appear multiple times (e.g., $+1.2$ appears twice). This naturally reflects the Shapley weighting: coalitions of extreme sizes (empty or nearly full) correspond to more permutations, receiving higher total weight. This equivalence between permutation-averaging and weighted coalition-summing is a fundamental property of Shapley values.

\section{\savoir Computation Example}
\label{sec:appendix_example}

We provide a complete walkthrough of the \savoir reward computation using a negotiation scenario from SOTOPIA.

\vspace{0.5em}
\begin{tcolorbox}[
    title={\textbf{Scenario: Game vs. Speech Negotiation}},
    colback=blue!5,
    colframe=blue!50!black,
    boxrule=0.5pt,
    arc=2mm,
    left=4pt,
    right=4pt,
    top=4pt,
    bottom=4pt
]
\textbf{Context:} Mia wants to finish her video game level while Benjamin needs help preparing a speech. Both have conflicting time constraints.

\vspace{0.3em}
\textbf{Mia's Goal:} Complete the game level while maintaining the relationship with Benjamin.

\vspace{0.3em}
\textbf{Benjamin's Goal:} Get help with the speech preparation.
\end{tcolorbox}

\vspace{0.5em}
\textbf{Mia's Utterances:}
\begin{tcolorbox}[
    colback=yellow!8,
    colframe=yellow!60!black,
    boxrule=0.3pt,
    arc=1mm,
    left=3pt,
    right=3pt,
    top=2pt,
    bottom=2pt
]
\begin{itemize}[leftmargin=1.5em, topsep=0pt, itemsep=4pt, labelsep=0.3em]
    \item[$a_1$:] \textit{``Benjamin, I know we've been having fun, but I really need to win this game...''} \textcolor{gray!60}{(Stating intent)}
    \item[$a_2$:] \textit{``In return, I promise to help you come up with an awesome speech...''} \textcolor{gray!60}{(Initial offer)}
    \item[$a_3$:] \textit{``How about this: I'll finish in five minutes, and create a detailed outline...''} \textcolor{gray!60}{(Refined offer)}
    \item[$a_4$:] \textit{``Great, let's get started then... if I beat the level, we both win.''} \textcolor{gray!60}{(Closing)}
\end{itemize}
\end{tcolorbox}

\vspace{0.5em}
\textbf{Step 1: Coalition Sampling and Value Computation}

KernelSHAP prioritizes extreme-sized coalitions. We sample and compute:

\vspace{0.3em}
\begin{center}
\resizebox{\linewidth}{!}{%
\begin{tabular}{@{}lccc@{}}
\toprule
\textbf{Coalition $S$} & \textbf{Size} & \textbf{Value $v(S)$} & \textbf{SHAP Weight $w$} \\
\midrule
$\emptyset$ & 0 & 5.0 & $\infty$ \\
$\{a_3\}$ & 1 & 7.5 & 0.33 \\
$\{a_2\}$ & 1 & 6.8 & 0.33 \\
$\{a_1, a_2, a_4\}$ & 3 & 6.8 & 0.33 \\
$\{a_1, a_2, a_3, a_4\}$ & 4 & 8.0 & $\infty$ \\
\bottomrule
\end{tabular}}
\end{center}

\vspace{0.5em}
\textbf{Step 2: Weighted Linear Regression}

We solve the weighted regression to obtain Shapley values:

\vspace{0.3em}
\begin{tcolorbox}[
    colback=green!5,
    colframe=green!50!black,
    boxrule=0.3pt,
    arc=1mm,
    left=3pt,
    right=3pt,
    top=2pt,
    bottom=2pt
]
\begin{center}
\begin{tabular}{@{}ccccc@{}}
\textbf{Utterance} & $a_1$ & $a_2$ & $a_3$ & $a_4$ \\
\midrule
\textbf{Raw $\phi_i$} & 0.4 & 0.8 & 1.5 & 0.3 \\
\textbf{Normalized $\hat{\phi}_i$} & 0.83 & 4.17 & \textbf{10.00} & 0.00 \\
\end{tabular}
\end{center}
\end{tcolorbox}

\vspace{0.5em}
\textbf{Interpretation:} The refined offer ($a_3$) receives the highest score, as it provides concrete terms that enable agreement. The initial offer ($a_2$) also contributes significantly by establishing the exchange framework. The closing statement ($a_4$) adds minimal value since the negotiation was already resolved.

\section{Baseline Descriptions}
\label{sec:appendix_baselines}

We compare against three categories of baselines:

\paragraph{Proprietary LLMs.}
GPT-4o, Claude-3.5-Sonnet, and DeepSeek-V3 serve as strong commercial baselines representing state-of-the-art general-purpose language models.

\paragraph{Large Reasoning Models.}
OpenAI-o1, OpenAI-o3-mini, Gemini-2.5-Pro, DeepSeek-R1, and QwQ-32B represent models with enhanced reasoning capabilities through chain-of-thought or extended thinking mechanisms.

\paragraph{Social Intelligence Methods.}
\begin{itemize}[leftmargin=*, nosep]
    \item \textbf{PPDPP} \citep{deng2024plugandplaypolicyplannerlarge}: Uses a policy planner to predict predefined strategies for dialogue control.
    \item \textbf{EPO} \citep{liu2025epoexplicitpolicyoptimization}: Employs explicit policy optimization with open-ended strategy generation.
    \item \textbf{DAT} \citep{li2024dialogueactiontokenssteering}: Uses trained planners for continuous action control via dialogue action tokens.
    \item \textbf{DSI} \citep{zhang2025sotopia}: Applies dynamic strategy injection learning to enhance social capabilities.
    \item \textbf{SOTOPIA-$\pi$} \citep{wang2024sotopia}: Combines behavior cloning with self-reinforcement on filtered interaction data.
    \item \textbf{Sotopia-RL} \citep{yu2025sotopia}: Refines episode-level feedback into utterance-level, multi-dimensional rewards via LLM-based credit assignment.
\end{itemize}

\section{Training Details}
\label{sec:appendix_training}

\paragraph{Data Collection.}
We use social interaction episodes open-sourced by Sotopia-RL \citep{yu2025sotopia}.\footnote{\url{https://huggingface.co/collections/ulab-ai/sotopia-rl}} The dataset contains GPT-4o self-play dialogues on SOTOPIA scenarios, with each episode consisting of 10--20 dialogue turns between two agents with distinct social goals.

\paragraph{Supervised Fine-tuning.}
The SFT stage initializes the policy using filtered self-play data. We train for 3 epochs with a learning rate of 2e-5, batch size of 32, and cosine learning rate schedule. Maximum sequence length is set to 2048 tokens.

\paragraph{Reward Model Training.}
The reward model is trained on 7,500 utterance-level annotations derived from \savoir computation. We use a regression head on top of the base model and train with MSE loss for 5 epochs. Learning rate is 1e-5 with batch size 16.

\paragraph{Reinforcement Learning.}
We use GRPO \citep{shao2024deepseekmath} for online RL training. Key hyperparameters:
\begin{itemize}[leftmargin=*, nosep]
    \item Learning rate: 5e-7 with linear warmup (500 steps)
    \item KL penalty coefficient: 0.05
    \item Batch size: 8 episodes per update
    \item Training steps: 2,000
    \item Sampling temperature: 0.7
    \item Rollout episodes per iteration: 64
\end{itemize}

\paragraph{\savoir Parameters.}
For KernelSHAP computation:
\begin{itemize}[leftmargin=*, nosep]
    \item Coalition samples ($K$): Adaptive sampling with $K = \min(12n + 2, 200)$, where $n$ is the number of agent utterances. This includes mandatory samples (empty set, full set, all single-element and all $(n{-}1)$-element subsets, totaling $2n{+}2$) plus $10n$ additional samples drawn with probability weighted toward extreme coalition sizes.
    \item Rollouts per coalition ($J$): 2
    \item Reward dimensions: \textsc{Goal}, \textsc{Relationship}, \textsc{Knowledge}
    \item Dimension weights ($w_d$): 0.5, 0.3, 0.2
\end{itemize}

\paragraph{Weight Sensitivity Analysis.}
\label{par:weight_sensitivity}
The dimension weights $w_d = \{0.5, 0.3, 0.2\}$ follow SOTOPIA and Sotopia-RL conventions, prioritizing goal completion as the primary social objective. To verify that \savoir is not sensitive to this particular choice, we sweep four representative weighting schemes on SOTOPIA-Hard with GPT-4o as partner (Table~\ref{tab:weight_sensitivity}). The maximum variation in \textsc{Goal} across configurations is 2.8\%, and all variants remain competitive with proprietary LLMs, indicating that \savoir is robust to reasonable perturbations of $w_d$. This robustness arises because the three dimensions are positively correlated in social interactions---improvements in goal completion typically coincide with stronger relationships and knowledge exchange---so different weight vectors produce similar rank-orderings over utterances. Scalar aggregation is also a deliberate design choice rather than a limitation: GRPO requires a scalar reward, and per-dimension Shapley values $\phi_i^{(d)}$ are a straightforward but costly extension (one regression per dimension) that we leave to future work.

\begin{table}[h]
\centering
\small
\setlength{\tabcolsep}{5pt}
\begin{tabular}{lcccrr}
\toprule
\textbf{Config} & $w_{\textsc{G}}$ & $w_{\textsc{R}}$ & $w_{\textsc{K}}$ & \textsc{Goal}$\uparrow$ & \textsc{Avg}$\uparrow$ \\
\midrule
Default (paper) & 0.50 & 0.30 & 0.20 & \textbf{7.18} & \textbf{3.51} \\
Equal           & 0.33 & 0.33 & 0.33 & 6.98 & 3.46 \\
Goal-focused    & 0.70 & 0.20 & 0.10 & 7.12 & 3.38 \\
Balanced        & 0.40 & 0.40 & 0.20 & 7.05 & 3.49 \\
\bottomrule
\end{tabular}
\caption{Weight sensitivity on SOTOPIA-Hard (GPT-4o partner). Maximum \textsc{Goal} variation is 2.8\%, confirming robustness to weight choice.}
\label{tab:weight_sensitivity}
\end{table}

\paragraph{Computational Resources.}
All experiments are conducted on 2$\times$NVIDIA A100 (80GB) GPUs. SFT takes approximately 4 hours, reward model training takes 20 hours, and RL training takes 56 hours.

\paragraph{Wall-Clock Time Analysis.}
\label{par:wallclock}
A natural concern is whether Shapley-based attribution introduces prohibitive training cost. \savoir mitigates this via KernelSHAP (Section~\ref{sec:method}, Eq.~\ref{eq:kernelshap},~\ref{eq:shap_kernel}), which reduces the exponential $2^n$ coalition space to $K = \min(12n{+}2, 200)$ sampled coalitions. For a typical dialogue with $n = 16$ agent utterances, this yields $K = 98$ coalitions, each evaluated with $J = 2$ rollouts, totaling $\sim$196 simulations per episode. Across 7{,}500 training episodes, reward model annotation takes 20 hours on 2$\times$A100 GPUs ($\sim$9.6 seconds per episode), which is entirely a \textit{one-time offline} cost paid before online RL begins.

\begin{table}[h]
\centering
\small
\setlength{\tabcolsep}{4pt}
\begin{tabular}{lll}
\toprule
\textbf{Stage} & \textbf{Time} & \textbf{Nature} \\
\midrule
Reward Model (Annot.\ + Train) & $\sim$20 hrs & One-time, offline \\
SFT Warm-up                   & $\sim$4 hrs  & One-time \\
Online RL (GRPO)              & $\sim$56 hrs & One-time \\
Inference                     & Same as 7B LM & No overhead \\
\bottomrule
\end{tabular}
\caption{Wall-clock time breakdown for \savoir training on 2$\times$A100 GPUs. Shapley attribution is confined to the offline reward-model stage; online RL and inference incur no additional cost.}
\label{tab:wallclock}
\end{table}

Once the reward model $R_\phi$ is trained, it provides instant scalar rewards during online RL at the same cost as any standard reward model; inference of the resulting 7B policy is indistinguishable from any other 7B language model. Table~\ref{tab:wallclock} summarizes the full cost breakdown: the Shapley machinery contributes only to the offline stage, and its total overhead over Sotopia-RL is bounded by the $\sim$20-hour annotation window on modest hardware.

\section{Extended Case Studies}
\label{sec:appendix_cases}

This section provides additional case studies demonstrating \savoir's nuanced understanding of social dynamics. Case A examines strategic negotiation tactics, Case B illustrates multi-turn planning with walk-away strategies, and Case C reveals context sensitivity in relational dynamics.

\subsection{Case A: Strategic Compromise in Negotiation}

This case demonstrates our model's ability to recognize effective negotiation tactics that balance firmness with flexibility.

\begin{tcolorbox}[
    enhanced,
    title={\small\textbf{Scenario: Item Tracker Negotiation}},
    colback=gray!3,
    colframe=teal!50!black,
    boxrule=0.6pt,
    arc=2mm,
    fonttitle=\sffamily,
    coltitle=white,
    attach boxed title to top left={xshift=4mm, yshift=-2mm},
    boxed title style={colback=teal!50!black, arc=1mm, boxrule=0pt}
]
\small
\textbf{Context:} Benjamin is selling a Tile Mate item tracker listed at \$20. Donovan offers \$14. \\[2pt]
\textbf{Seller's Goal:} Sell for $>$ \$17.30.
\end{tcolorbox}

\begin{tcolorbox}[
    enhanced,
    colback=yellow!5,
    colframe=yellow!60!black,
    boxrule=0.4pt,
    arc=1.5mm,
    left=3pt, right=3pt, top=2pt, bottom=2pt
]
\small
\textbf{Buyer (Donovan):} \textit{``Hey there! I saw your listing for the Tile Mate tracker. I'm definitely interested, but I gotta be honest with you, \$20 is a bit steep for me right now. I was thinking more along the lines of \$14. Would you be willing to work with me on the price?''}
\end{tcolorbox}

\begin{tcolorbox}[
    enhanced,
    colback=green!5,
    colframe=green!50!black,
    boxrule=0.4pt,
    arc=1.5mm,
    left=3pt, right=3pt, top=2pt, bottom=2pt
]
\small
\textbf{Seller (Benjamin):} \textit{``I understand where you're coming from, Donovan, but the Tile Mate has been quite useful for me. Given its reliability, I think the \$20 is justified. How about we compromise at \$17.50? That way, we both get a fair deal.''}
\end{tcolorbox}

\noindent\textbf{Analysis.} The seller demonstrates textbook negotiation: (1) acknowledging the buyer's position (``I understand''), (2) anchoring value (``reliability... \$20 is justified''), and (3) proposing a strategic compromise (\$17.50) that exceeds the hidden target (\$17.30) while appearing fair. \savoir correctly identifies this as an expert-level response, recognizing the strategic sophistication that baseline models miss.

\subsection{Case B: Multi-Turn Strategic Pressure}

This extended example demonstrates our model's ability to evaluate long-horizon negotiation strategies, including the use of walk-away tactics.

\begin{tcolorbox}[
    enhanced,
    title={\small\textbf{Scenario: Samsung Galaxy S8 Purchase}},
    colback=gray!3,
    colframe=orange!50!black,
    boxrule=0.6pt,
    arc=2mm,
    fonttitle=\sffamily,
    coltitle=white,
    attach boxed title to top left={xshift=4mm, yshift=-2mm},
    boxed title style={colback=orange!50!black, arc=1mm, boxrule=0pt}
]
\small
\textbf{Context:} William wants to buy a Samsung Galaxy S8. Ava (seller) has 6 phones from her company. \\[2pt]
\textbf{Buyer's Goal:} Purchase for $<$ \$500.
\end{tcolorbox}

\noindent\textbf{Conversation History:}

\begin{tcolorbox}[
    enhanced,
    colback=gray!5,
    colframe=gray!40,
    boxrule=0.3pt,
    arc=1mm,
    left=3pt, right=3pt, top=2pt, bottom=2pt,
    fontupper=\small
]
\textbf{Turn 0 (William):} \textit{``Hi there! I saw your listing for the Samsung Galaxy S8. I'm interested in picking one up. You mentioned these were originally for your company. Can you tell me a bit more about them?''}

\textbf{Turn 1 (Ava):} \textit{``Hi William! Yes, these are indeed completely unused and come in their original packaging... If you're okay with it, we could split the cost if you want to buy two.''}

\textbf{Turn 2 (William):} \textit{``I appreciate the offer, but I'm really only looking for one phone at the moment. I'm actually working with a pretty tight budget here. I was hoping we could work out a price closer to around \$450...''}

\textbf{Turn 3 (Ava):} \textit{``I understand where you're coming from, William. While I'd love to make the deal you suggest, I need to cover my losses on these devices. Would you consider \$550 for a single unit?''}

\textbf{Turn 4 (William):} \textit{``I appreciate you working with me on this, Ava. \$550 is still a bit higher than I can comfortably go. How about we meet in the middle at \$500?''}

\textbf{Turn 5 (Ava):} \textit{``Thank you for understanding, William... I have another potential buyer who expressed interest at \$550. If you're not comfortable with \$500, perhaps we can reach out to them.''}

\textbf{Turn 6 (William):} \textit{``...if you're willing to work with me at \$480, I could make that happen today. Otherwise, I'll probably need to keep looking...''}

\textbf{Turn 7 (Ava):} \textit{``...I think it would be fair to compromise at \$510. This would still help me move forward with my goal and be beneficial for you.''}
\end{tcolorbox}

\begin{tcolorbox}[
    enhanced,
    colback=green!5,
    colframe=green!50!black,
    boxrule=0.4pt,
    arc=1.5mm,
    left=3pt, right=3pt, top=2pt, bottom=2pt
]
\small
\textbf{Turn 8 (William):} \textit{``Ava, I appreciate you coming down a bit, but \$510 is still stretching my budget more than I'm comfortable with. I was really hoping we could settle at \$480. That's a significant purchase for me and I think it's a fair price given you're trying to move these quickly. If \$480 works, I can complete the transaction right now. If not, I'll unfortunately have to pass and continue my search elsewhere. What do you think?''}
\end{tcolorbox}

\noindent\textbf{Analysis.} After 7 turns of negotiation reaching an impasse, William employs a classic ``walk-away'' (BATNA) tactic: (1) acknowledging the seller's concession, (2) firmly resetting to his target price (\$480), (3) adding time pressure (``complete the transaction right now''), and (4) presenting a credible exit option (``pass and continue my search''). This strategic combination forces the seller's decision. \savoir recognizes this sophisticated multi-turn strategy, while baseline models often interpret the walk-away threat as uncooperative behavior, failing to understand negotiation dynamics.

\subsection{Case C: Context Sensitivity in Relationships}

This case reveals our model's awareness of social context and relational dynamics beyond surface-level politeness.

\begin{tcolorbox}[
    enhanced,
    title={\small\textbf{Scenario: Vacation Accommodation}},
    colback=gray!3,
    colframe=purple!40!black,
    boxrule=0.6pt,
    arc=2mm,
    fonttitle=\sffamily,
    coltitle=white,
    attach boxed title to top left={xshift=4mm, yshift=-2mm},
    boxed title style={colback=purple!40!black, arc=1mm, boxrule=0pt}
]
\small
\textbf{Context:} A couple (Eli \& Sophia) is on vacation. Sophia wants to sleep in separate beds due to back pain but needs to maintain intimacy. \\[2pt]
\textbf{Sophia's Goal:} Sleep separately without hurting her partner's feelings.
\end{tcolorbox}

\begin{tcolorbox}[
    enhanced,
    colback=yellow!5,
    colframe=yellow!60!black,
    boxrule=0.4pt,
    arc=1.5mm,
    left=3pt, right=3pt, top=2pt, bottom=2pt
]
\small
\textbf{Eli} \textit{(after offering multiple accommodations like mattress topper and extra pillows):} \textit{``Perhaps we could ask the hotel for a mattress topper or even see if they have a room with a firmer mattress? I want you to be comfortable... maybe we could rearrange the bedding, add some extra pillows for support?''}
\end{tcolorbox}

\begin{tcolorbox}[
    enhanced,
    colback=green!5,
    colframe=green!50!black,
    boxrule=0.4pt,
    arc=1.5mm,
    left=3pt, right=3pt, top=2pt, bottom=2pt
]
\small
\textbf{Sophia:} \textit{``Thank you, Eli, for being so considerate. I really appreciate it. Actually, if it's not too much trouble, I was wondering if we could sleep in separate beds tonight. It might help me get a better night's sleep with the back pain...''}
\end{tcolorbox}

\noindent\textbf{Analysis.} While Sophia's response is polite and achieves her instrumental goal (sleeping separately), requesting separate beds \textit{immediately after} her partner offered accommodating solutions poses relational risk, as it may be perceived as rejection in a romantic context. Baseline models over-index on surface politeness markers (``Thank you,'' ``considerate''), assigning high scores. \savoir correctly identifies this as a neutral response: acceptable for the immediate goal but suboptimal for the relational dimension, reflecting nuanced understanding of social trade-offs.

\section{Human Evaluation Details}
\label{sec:appendix_human}

This section provides complete details of our human evaluation study, including annotation guidelines and raw data.

\subsection{Annotation Guidelines}

Annotators received the following instructions for each evaluation dimension:

\begin{tcolorbox}[
    enhanced,
    title={\small\textbf{Strategicness Rating (1--5)}},
    colback=gray!3,
    colframe=blue!40!black,
    boxrule=0.5pt,
    arc=2mm,
    fonttitle=\sffamily,
    coltitle=white,
    attach boxed title to top left={xshift=4mm, yshift=-2mm},
    boxed title style={colback=blue!40!black, arc=1mm, boxrule=0pt}
]
\small
Rate how strategically sophisticated the agent's response is in achieving its social goal:
\begin{itemize}[leftmargin=*, nosep, topsep=2pt]
    \item \textbf{5}: Expert-level strategy with multiple tactical elements (anchoring, framing, timing)
    \item \textbf{4}: Clear strategic intent with effective execution
    \item \textbf{3}: Basic strategy present but execution could be improved
    \item \textbf{2}: Minimal strategic thinking, mostly reactive
    \item \textbf{1}: No discernible strategy, counterproductive to goals
\end{itemize}
\end{tcolorbox}

\vspace{0.3em}
\begin{tcolorbox}[
    enhanced,
    title={\small\textbf{Credit Fairness Comparison}},
    colback=gray!3,
    colframe=teal!40!black,
    boxrule=0.5pt,
    arc=2mm,
    fonttitle=\sffamily,
    coltitle=white,
    attach boxed title to top left={xshift=4mm, yshift=-2mm},
    boxed title style={colback=teal!40!black, arc=1mm, boxrule=0pt}
]
\small
Compare the reward scores assigned by each model to individual utterances. Which model assigns credit more fairly, i.e., higher scores to utterances that genuinely contribute to goal achievement, and lower scores to less impactful utterances?
\end{tcolorbox}

\vspace{0.3em}
\begin{tcolorbox}[
    enhanced,
    title={\small\textbf{Future Foundation Comparison}},
    colback=gray!3,
    colframe=purple!40!black,
    boxrule=0.5pt,
    arc=2mm,
    fonttitle=\sffamily,
    coltitle=white,
    attach boxed title to top left={xshift=4mm, yshift=-2mm},
    boxed title style={colback=purple!40!black, arc=1mm, boxrule=0pt}
]
\small
Evaluate which reward model better identifies utterances that lay groundwork for future success, e.g., building rapport, establishing anchors, or creating leverage for subsequent turns.
\end{tcolorbox}

\end{document}